\documentclass[11pt,onecolumn,a4paper]{style}

\usepackage{longtable}
\usetikzlibrary{fit}

\newcommand{\nocitehere}{\renewcommand{\cite}[2][]{}}
\makeatletter
\patchcmd{\LT@output}{\vss}{\vskip\z@\@plus1fil\@minus\normalbaselineskip}{}{}
\patchcmd{\LT@output}{\vss}{\vskip\z@\@plus1fil\@minus\normalbaselineskip}{}{}
\makeatother

\title{HIVE-Bench: Evaluating Patch-Level Visual \\ Representations for Egocentric Robot Manipulation}

\author{%
\begin{HIVEauthorblock}
\HIVEauth{Qiwei Liang}{1,2}{*,\S}\authsep
\HIVEauth{Zhengtu Liang}{1}{*}\authsep
\HIVEauth{Minghao Lai}{3,4}{*}\authsep
\HIVEauth{Yikeng Chen}{1}{*}\authsep
\HIVEauthAff{Yiming Wang}{5,2}\authsep
\HIVEauthAff{Boyang Cai}{1}\authbreak
\HIVEauthAff{Yuetong Fang}{1}\authsep
\HIVEauthAff{Taowen Wang}{1}\authsep
\HIVEauthAff{Baiqiao Yin}{6}\authsep
\HIVEauthAff{Tianxing Chen}{7}\authsep
\HIVEauthAff{Yue Chen}{8}\authsep
\HIVEauthAff{Jiaming Liang}{1}\authbreak
\HIVEauthAff{Guangyu Chen}{9}\authsep
\HIVEauthAff{Shaolong Zhu}{9}\authsep
\HIVEauth{Daolin Ma}{5,2}{\ddag}\authsep
\HIVEauth{Renjing Xu}{1}{\ddag}
\end{HIVEauthorblock}
}

\affiliation{%
\begin{HIVEaffiliationblock}
\affitem{1}{The Hong Kong University of Science and Technology (Guangzhou)} \quad
\affitem{2}{Xense Robotics}\\[0.15em]
\affitem{3}{Institute of Automation, Chinese Academy of Sciences} \quad
\affitem{4}{Zhongguancun Academy} \quad
\affitem{5}{Shanghai Jiao Tong University}\\[0.15em]
\affitem{6}{Nanyang Technological University} \quad
\affitem{7}{The University of Hong Kong} \quad
\affitem{8}{Peking University} \quad
\affitem{9}{Tsinghua University}
\end{HIVEaffiliationblock}
}

\correspondence{%
  \metamark{*} Equal contribution \HIVEmetasep
  \metamark{\S} Project leader \HIVEmetasep
  \metamark{\ddag} Corresponding author
}

\date{}

\begin{document}
\maketitle
\setlength{\parfillskip}{0pt plus 0.5\linewidth}
\makeatletter
\apptocmd{\@parboxrestore}{\parfillskip=0pt plus 0.5\linewidth\relax}{}{}
\makeatother

\begin{abstract}
Pretrained vision is a common start for robot learning, but most comparisons still score one pooled vector from a third-person image. Bimanual policies instead read dense patch tokens from cameras on the robot. We introduce \textbf{HIVE-Bench}: one shared flow-matching policy reads 20 encoders from seven families on 24 egocentric bimanual tasks, across adaptation, task sharing, and sweeps of data, model size, and layer. Keeping the patches matters. Averaging each view to one token removes most of the success, adaptation helps every pair we test, and one shared policy keeps most of the success of separate policies at a fraction of the updates. The same tokens are then scored with statistics, state and object readouts, published vision scores, and an inverse-dynamics readout. That readout of the commanded motion is one of the strongest measurements available before a policy is trained. Recognition, depth, and correspondence do not transfer reliably. Fine-tuning can improve the action readout even when other measurements move the other way, and no statistic is stable enough to replace a rollout before any deployment \mbox{\href{https://github.com/KolaKivy/HIVE-Bench}{\textbf{Codebase \& Benchmark Zoo.}}}

\end{abstract}
\par
\vspace{12pt}
\noindent\begin{minipage}{\linewidth}
\centering
\captionsetup{hypcap=false,skip=3pt}
\includegraphics[width=\linewidth]{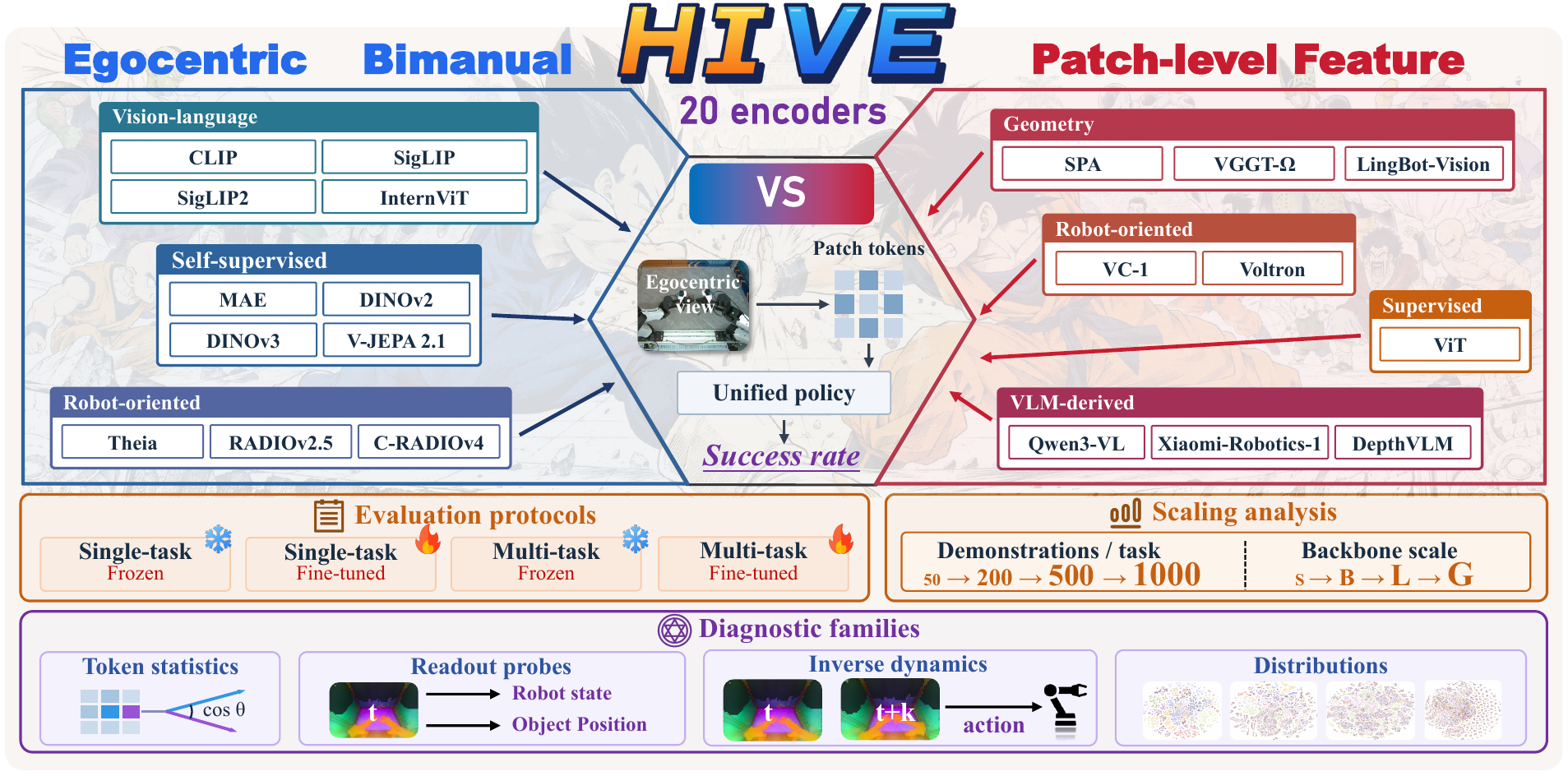}
\captionof{figure}{\textbf{HIVE-Bench.} Twenty pretrained encoders from seven families compete through one shared patch-conditioned policy on egocentric bimanual tasks, across data and model scales, training regimes, and adaptation.}
\label{fig:teaser}
\end{minipage}
\clearpage

\section{Introduction}
\label{sec:m-intro}

Robot manipulation policies increasingly reuse pretrained visual encoders instead of learning vision from scratch, which transfers poorly across objects, robots, and viewpoints~\citep{r3m,cortex,mcr,parisi,generalizationgap}. The candidates come from very different signals, including self-supervision, language alignment, 3D reconstruction, robot data, and vision--language models~\citep{mae,dinov2,clip,spa,qwen3}. Choosing among them still means training and rolling out a policy. A benchmark with a shared policy can reveal which representations transfer, and whether inexpensive measurements explain their successes and failures across different training conditions.

Earlier evaluations often read a pooled third-person feature and relate scene reconstruction to control~\citep{cortex,hucontrol,datasets,burns,theia,mcr,structure}. Bimanual policies do not use that interface. They attend to dense patch tokens, which keep the local detail that separates nearby grasps~\citep{openvla,pi0,patchpolicy,afa}, and they see through head and wrist cameras that move with the robot and are often occluded~\citep{act,mobilealoha,robotwin,umi}. We therefore ask how encoders compare when one policy reads their patch tokens from onboard cameras, and which properties of those tokens track the resulting success.

We introduce \textbf{HIVE-Bench} (Figure~\ref{fig:teaser}), to our knowledge the first comprehensive benchmark of patch-level representations for egocentric bimanual manipulation that jointly studies control and diagnostic correlations across training regimes. A flow-matching action head~\citep{dit,gr00t} reads 20 encoders from seven families on two egocentric bimanual suites~\citep{robocasa,robotwin}. We train it frozen and fine-tuned, on one task and on a shared suite, and we sweep data, model size, and the extracted layer, because a ranking from a single regime is easy to over-read. The same tokens are then scored in several ways: token statistics, state and object readouts, published vision benchmarks, and an inverse-dynamics readout (IDM) that recovers the commanded motion.

Those runs support three simple conclusions. Keeping the patch tokens, rather than averaging each view into one, accounts for most of the success of the encoders we compare this way. Fine-tuning improves every encoder we adapt, and one policy shared across tasks keeps most of the success of separate policies at a fraction of the updates. The measurements are the second half of the study. Several of them correlate with success somewhere, but none is stable enough to prove which encoder a robot should use. IDM is among the strongest readouts, especially on RoboTwin, yet even its ranking misses the best encoder and fails to predict gains from adaptation. Our goal is to map these relationships and their limits; finding measurements that improve on IDM remains an open research problem. Our contributions are:
\begin{itemize}
\item \textbf{An egocentric patch-level benchmark:} one shared policy reads the dense patch tokens, not a pooled vector, of 20 encoders from onboard cameras on two bimanual suites under one shared update budget, through documented token interfaces (Section~\ref{sec:m-design}).
\item \textbf{Evaluation from many angles:} single-task and multi-task training, frozen and fine-tuned encoders, demonstration and model-size scaling, extraction layers, pooling, and optional robot state, with the action head and optimizer held fixed (Section~\ref{sec:m-results}).
\item \textbf{A study of what tracks success, and where it stops:} token statistics, action and state readouts, and published vision scores, compared across suites, task sharing, and adaptation, with the cases where a frozen ranking fails to predict improvement kept separate from the cases where it holds, and with IDM among the strongest metrics (Sections~\ref{sec:m-diag} and~\ref{sec:m-disc}).
\end{itemize}

\section{Related Work}
\label{sec:m-related}

\paragraph{Pretrained representations for control.}
Global embeddings made pretrained vision easy to reuse in control~\citep{parisi,r3m,cortex}. Later encoders learn from robot data or distill several teachers~\citep{voltron,spa,theia,radiov25,radio,aem,afro}. Evaluations have compared adaptation, downstream learners, data, and objectives~\citep{cortex,hucontrol,datasets,explorepretrain}, usually from one pooled third-person feature and a separately trained policy. Newer studies keep more spatial input, but still differ in embodiment, task sharing, and what they measure (Table~\ref{tab:m-coverage}).

\paragraph{Spatial policy interfaces and egocentric bimanual manipulation.}
ALOHA, ACT, Mobile ALOHA, UMI, DexMimicGen, and GR00T advanced bimanual skills with onboard views, dexterous hands, and humanoids~\citep{act,mobilealoha,umi,dexmimicgen,gr00t,dqnet,mmabc}. Policies differ in how much spatial detail they keep, from TokenLearner compression in RT-1 to dense pretrained tokens in Patch Policy~\citep{rt1,octo,patchpolicy}, and VLA systems and world-action models integrate vision with action prediction~\citep{rt2,openvla,pi0,pi05,dreamzero,fastwam,liang2026univep}. HIVE-Bench holds the action head fixed, so only the encoder and its token interface change between two compared policies.

\paragraph{Diagnosing representation quality.}
Proposed proxies include segmentation ability~\citep{burns}, objectness~\citep{slotmim}, manipulation relevance~\citep{mcr}, token-norm entropy~\citep{theia}, attention allocation~\citep{ver}, effective rank~\citep{rankme}, physical structure~\citep{spa,structure}, temporal progression~\citep{temporaltrap}, and action recoverability from world-model latents~\citep{semanticwm}. Each measures one aspect under its own policy. We measure token statistics, state and action readouts~\citep{spatialprobe}, and published vision scores under one patch-level policy, across task sharing, scale, and adaptation, and test which associations survive a change of setting. Table~\ref{tab:m-coverage} places this coverage next to the closest prior studies.

\begin{center}
\vspace{-2pt}
\captionof{table}{\textbf{Coverage of representation evaluation.} Ego: onboard views. Bi: bimanual. Patch: spatial tokens. F/FT: frozen and fine-tuned. S/M: single- and multi-task. Stat, State, and Act: token statistics, state decoding, and action readout. Corr: association with success.}
\label{tab:m-coverage}
\begingroup\scriptsize\setlength{\tabcolsep}{3pt}\renewcommand{\arraystretch}{0.96}
\newcommand{\covyes}{\textcolor{hiveTeal}{\(\boldsymbol{\checkmark}\)}}
\newcommand{\covno}{\textcolor{hiveRose}{\(\times\)}}
\begin{tabular*}{\linewidth}{@{\extracolsep{\fill}}l*{9}{c}@{}}
\toprule
\textbf{Study} & \multicolumn{3}{c}{\textbf{Setting}} & \multicolumn{2}{c}{\textbf{Training}} & \multicolumn{4}{c}{\textbf{Diagnostics}}\\
\cmidrule(lr){2-4}\cmidrule(lr){5-6}\cmidrule(lr){7-10}
& Ego & Bi & Patch & F/FT & S/M & Stat & State & Act & Corr\\
\midrule
VC-1 (2023)~\cite{cortex} & \covyes & \covno & \covyes & \covyes & \covno & \covno & \covno & \covno & \covno\\
Hu et al. (2023)~\cite{hucontrol} & \covno & \covno & \covno & \covno & \covno & \covno & \covyes & \covno & \covyes\\
Burns et al. (2023)~\cite{burns} & \covyes & \covyes & \covyes & \covno & \covno & \covno & \covno & \covno & \covyes\\
Theia (2024)~\cite{theia} & \covyes & \covno & \covyes & \covyes & \covno & \covyes & \covno & \covno & \covyes\\
SPA (2025)~\cite{spa} & \covyes & \covyes & \covyes & \covno & \covyes & \covno & \covyes & \covno & \covyes\\
CVESC (2026)~\cite{structure} & \covno & \covno & \covno & \covyes & \covno & \covno & \covyes & \covno & \covyes\\
RoS (2026)~\cite{semanticwm} & \covno & \covno & \covyes & \covno & \covno & \covno & \covno & \covyes & \covno\\
Patch Policy (2026)~\cite{patchpolicy} & \covyes & \covno & \covyes & \covno & \covyes & \covno & \covno & \covno & \covno\\
\midrule
\textbf{HIVE-Bench} & \covyes & \covyes & \covyes & \covyes & \covyes & \covyes & \covyes & \covyes & \covyes\\
\bottomrule
\end{tabular*}
\endgroup
\vspace{-4pt}
\end{center}

\section{HIVE-Bench}
\label{sec:m-design}

HIVE-Bench compares encoders by the closed-loop success of one shared patch-conditioned policy, and asks which measurable properties of the tokens go with that success. The environments that supply the tasks and the policy that reads them are introduced separately below. Appendix~\ref{sec:design} gives the implementation and the training recipe.

\subsection{Environments and protocols}
\label{sec:m-bench}

\begin{figure}[t]
\centering
\includegraphics[width=\linewidth]{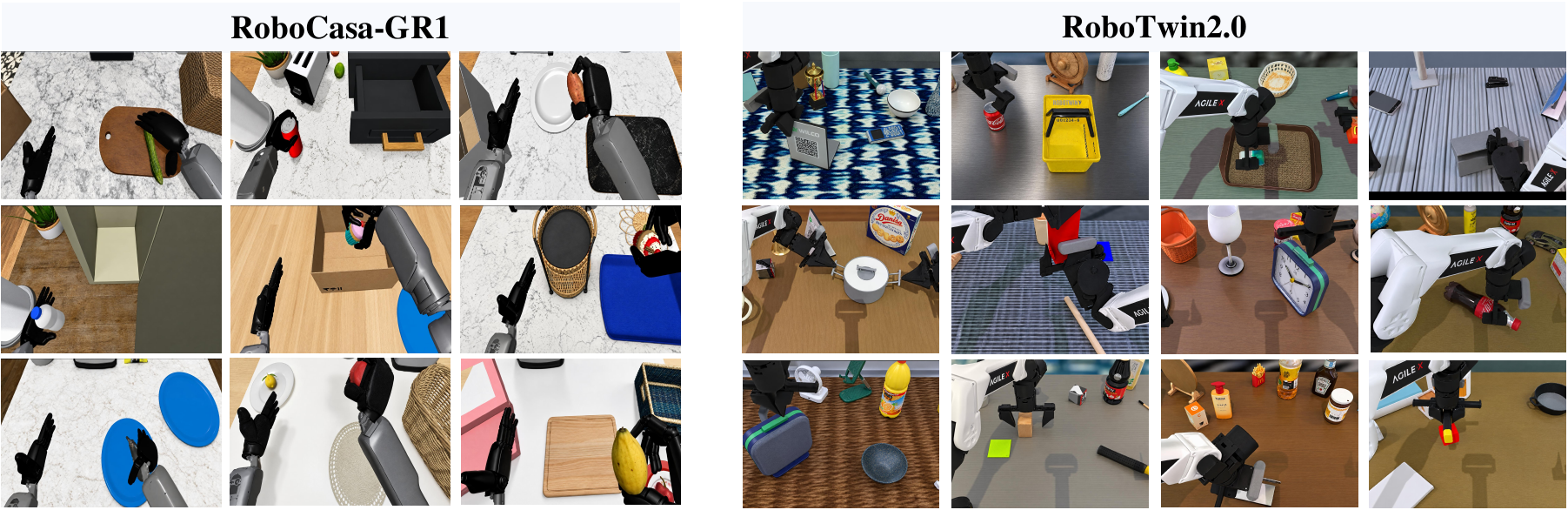}
\caption{\textbf{Environments.} Sample scenes from RoboCasa-GR1 (left), where a humanoid with dexterous hands works in varied kitchens, and RoboTwin~2.0 (right), where two arms with grippers work on tabletops with randomized clutter, lighting, and textures.}
\label{fig:m-simulation}
\end{figure}

We use 12 RoboCasa-GR1 kitchen tasks, performed by a humanoid whose 29-D action adds difficulty~\citep{robocasa,gr00t,dexac}, and 12 domain-randomized RoboTwin~2.0 tasks, performed by two arms with grippers~\citep{robotwin,robonvs} (Figure~\ref{fig:m-simulation}; Table~\ref{tab:app-tasks}). RoboCasa covers object transfer and cabinet, drawer, and microwave interactions. RoboTwin covers tool use, handover, placement, and articulated objects under randomized clutter, lighting, and textures. The suites differ in robot, action, cameras, tasks, and data at once, so differences between them are observational. Default policies see only the current images. One ablation adds robot state (Section~\ref{sec:m-inputs}).

\begin{table}[t]
\caption{\textbf{Environments and protocols.} Both have 12 tasks and predict 16-step action chunks. Demos are per task. ``Executed'' is the part of each chunk run before replanning. The last columns give the training regime and the number of encoders. Default policies use images without proprioception; diagnostics read the head camera.}
\label{tab:m-bench}
\centering\footnotesize
\setlength{\tabcolsep}{2pt}
\begin{tabular*}{\linewidth}{@{\extracolsep{\fill}}llcccccc@{}}
\toprule
\textbf{Benchmark} & \textbf{Robot} & \textbf{Demos} & \textbf{Views} & \textbf{Action} & \textbf{Executed} & \textbf{Frozen} & \textbf{Fine-tuned}\\
\midrule
RoboCasa-GR1 & Humanoid, hands & 1{,}000 & Head & 29-D & 12 of 16 & 17 ST, 20 MT & 5 ST\\
RoboTwin 2.0 & Two arms & 500 & Head, 2 wrists & 14-D joints & 16 of 16 & 20 MT & 5 MT\\
\bottomrule
\end{tabular*}
\end{table}

Multi-task training (MT) fits one language-conditioned policy to the twelve tasks of a suite. Single-task training (ST) fits a separate policy to each RoboCasa task. The closed-loop comparison covers 20 frozen encoders under MT, 17 of them under ST, and five fine-tuned encoders on RoboCasa ST and on RoboTwin MT (Table~\ref{tab:m-bench}). Further controls vary the number of demonstrations, the checkpoint size, the extraction layer, mean pooling, the state input, and the learned queries of the action head. Every configuration in the main comparison is trained with one seed and evaluated with three evaluation seeds of 50 rollouts per task. A suite score is the mean of the twelve task scores, and its deviation propagates the per-task evaluation deviations. A three-seed RoboCasa MT study, including its stability analysis, is reported in Appendix~\ref{sec:results-seeds}.

\subsection{Encoders and the token interface}
\label{sec:m-encoders}

The 20 encoders span seven families (Table~\ref{tab:encoder_families}): a supervised ViT~\citep{vit}; the self-supervised MAE, DINOv2, DINOv3, and V-JEPA~2.1~\citep{mae,dinov2,dinov3,vjepa21}; the geometry-oriented SPA, VGGT-$\Omega$, and LingBot-Vision~\citep{spa,vggtomega,lingbot}; the vision--language CLIP, SigLIP, SigLIP2, and InternViT~\citep{clip,siglip,siglip2,internvl}; the robot-oriented VC-1 and Voltron~\citep{cortex,voltron}; the multi-teacher Theia, RADIOv2.5, and C-RADIOv4~\citep{theia,radiov25,radio}; and three VLMs, Qwen3-VL and its derivatives DepthVLM and Xiaomi-Robotics-1~\citep{qwen3,depthvlm,xiaomi}. Table~\ref{tab:suite} lists sizes and token counts. Appendix~\ref{app:encoders} documents checkpoints and preprocessing.

The policy reads every patch token of every view. Class and register tokens are removed, and there is no extra feature normalization at the policy interface. Views are stored at $224\times224$, giving 196 or 256 tokens per view for patch size 16 or 14. V-JEPA~2.1 resizes to 384 pixels and produces 576 tokens, and each VLM contributes 64 final-layer image tokens. Those VLM states do not depend on the instruction, and we do not use Xiaomi-Robotics-1's action head. Two interfaces mix views before the policy reads them: VGGT-$\Omega$ receives RoboTwin's three views as one input, and a VLM's image tokens of later views can attend to earlier ones. We therefore compare an encoder under a documented extraction interface, not in isolation (Appendix~\ref{app:encoders}).

\subsection{The shared policy}
\label{sec:m-policy}

\begin{figure}[t]
\centering
\includegraphics[width=\linewidth]{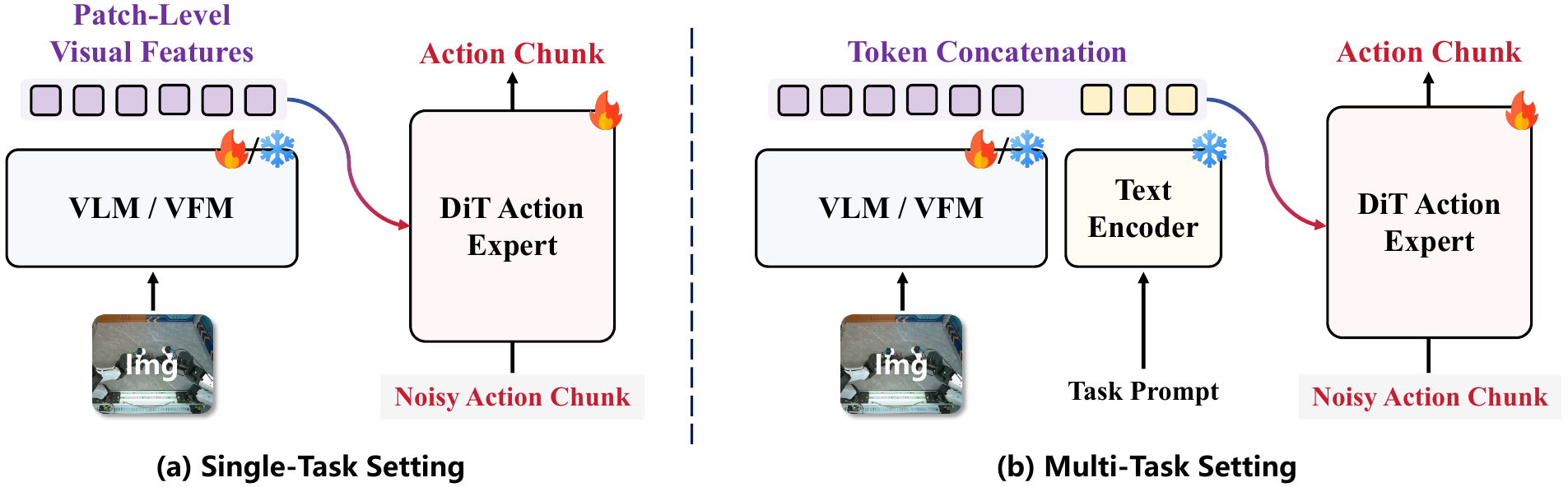}
\caption{\textbf{The shared policy.} (a) Single-task: patch tokens from a frozen or fine-tuned encoder, a vision foundation model or a VLM, condition a DiT action expert that denoises a 16-step action chunk. (b) Multi-task: visual and frozen text tokens condition the action expert, with task instructions supplied alongside the visual observations. Flames mark trained modules and snowflakes mark frozen ones.}
\label{fig:m-policy}
\end{figure}

The action head is a DiT flow-matching model derived from GR00T~\citep{gr00t,dit,flow}, with 16 blocks of width 768 and twelve heads of dimension $d_h=64$ (Figure~\ref{fig:m-policy}). It operates on 32 learned query tokens and 16 noisy action tokens $H_t$. The queries are latent positions for attention, and only the action positions are decoded into a velocity. They are distinct from the encoder's register tokens, which are discarded before this head sees the image. Eight blocks cross-attend to the context $C_t$ and eight apply self-attention, in alternation, and adaptive layer normalization conditions every block on the flow time. $C_t$ concatenates the patch tokens of all views. Multi-task runs append projected features of a frozen CLIP text encoder alongside visual tokens for task conditioning, so vision enters the policy only through
\begin{equation}
\mathrm{CrossAttn}(H_t,C_t)=\mathrm{softmax}\!\big((H_tW_Q)(C_tW_K)^\top/\sqrt{d_h}\big)\,C_tW_V .
\label{eq:m-xattn}
\end{equation}
The head regresses the flow-matching velocity toward the normalized 16-step action chunk. The flow time is drawn from $\mathrm{Beta}(1.5,1)$, with eight noise draws per sample, and inference integrates 15 Euler steps on RoboCasa or 10 on RoboTwin. Action heads span 138.669--156.478M parameters. Appendix~\ref{sec:design} gives complete downstream counts and the flow objective.

Each policy is trained for 100k AdamW steps with a global batch of 128 on four GPUs, in BF16, with warmup and cosine decay. The budget is equal per policy, not per task, so twelve single-task policies consume twelve times the updates of one multi-task policy on the same suite. The action head uses a peak learning rate of $10^{-4}$. Fine-tuning updates the encoder at $10^{-5}$. Other trainable projections use $3\times10^{-5}$ on RoboCasa MT and $10^{-4}$ otherwise. Frozen runs update the head and its projections only. Table~\ref{tab:suite} reports training FLOPs and step time, and Appendix~\ref{app:protocol} gives the complete recipe.

\subsection{Diagnostics and statistics}
\label{sec:m-diagdefs}

The diagnostics describe each encoder from head-camera frames, before asking which of those descriptions agree with closed-loop success (Appendix~\ref{app:metrics}).
\begin{itemize}
\item \textbf{Token statistics} need only frames: patch similarity, norms, covariance rank, spatial frequency, and their change through an episode. \emph{Norm-histogram entropy}~\citep{theia} bins each frame's patch norms into 30 bins. Features come from the early frames of every fifth demonstration.
\item \textbf{IDM and state probes} use four-layer, width-256 transformers and a learned query; only input projections grow with encoder width. The \emph{IDM probe} reads frames 16 steps apart and predicts the standardized command displacement $a_{t+15}-s_t$. A single-frame control predicts that same target from $I_t$, and the \emph{state probe} decodes $s_t$. An object probe covers four RoboTwin tasks. Because the IDM target contains $s_t$, part of it is readable from one frame (Appendix~\ref{app:probes}).
\item \textbf{Published vision scores} are collected for the checkpoints that report them under a shared protocol, covering classification, segmentation, depth, correspondence, and label propagation (Section~\ref{sec:m-vision}).
\end{itemize}

Associations use Spearman's $\rho$, a permutation test, a 95\% bootstrap interval, and a Benjamini--Hochberg correction within each cohort. Primary cohorts are the vision encoders; readouts also include the VLMs. Family deletion and paired before--after comparisons test whether an association depends on related encoders or on adaptation (Appendix~\ref{app:stats}).

These measurements are cheaper than training the policy. Token statistics need only demonstration frames, the probes train for four epochs on frozen tokens, and the vision scores come from published evaluations rather than new robot runs. Section~\ref{sec:m-diag} then asks which of them agrees with closed-loop success, and which of those agreements disappear when the benchmark or the training regime changes.

\section{What Makes a Visual Representation Useful for Control?}
\label{sec:m-results}

We first examine the policy interface, then encoder choice, task sharing, scale, and additional inputs. Success rates are percentages; changes are in percentage points.

\subsection{Preserve spatial tokens, then adapt the encoder}

\begin{figure}[t]
\centering
\includegraphics[width=\linewidth]{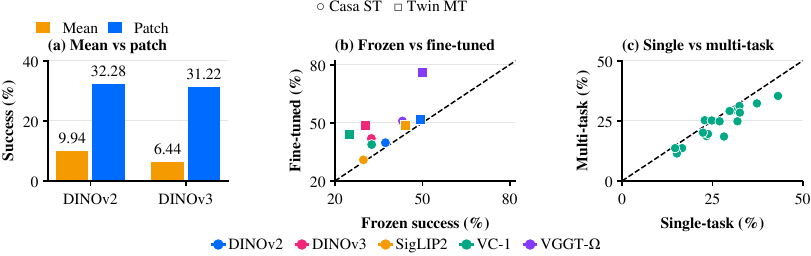}
\caption{\textbf{Spatial tokens, adaptation, and task sharing.} (a) Mean pooling versus patch tokens on RoboCasa MT. (b) Frozen versus fine-tuned success on RoboCasa ST (circles) and RoboTwin MT (squares). (c) RoboCasa ST versus MT, with 100k training steps per policy and the same demonstrations per task. Dashed lines denote equal success.}
\label{fig:m-closedloop}
\end{figure}

\paragraph{\textcolor{hiveAccent}{Mean pooling discards much of the information the policy uses.}}\label{fnd:gap}
Replacing each view's patches with their mean and retraining the same head reduces DINOv2 from 32.28\% to 9.94\% and DINOv3 from 31.22\% to 6.44\% on RoboCasa MT (Figure~\ref{fig:m-closedloop}a). The loss appears in 23 of 24 encoder--task pairs (Appendix~\ref{sec:results-gap}). Scene identity can survive pooling while local distinctions needed to act disappear; evaluating individual patches exposes that difference.

\paragraph{\textcolor{hiveAccent}{Fine-tuning improves success, with returns that depend on the encoder.}}\label{fnd:ft-headroom}
Jointly adapting five encoders and the head improves all ten benchmark pairs (Figure~\ref{fig:m-closedloop}b): gains span 1.11--9.44 points on RoboCasa ST and 2.22--26.06 on RoboTwin MT. VGGT-$\Omega$ rises from 50.00\% to 76.06\% on RoboTwin, whereas DINOv2 moves from 49.39\% to 51.61\%. The five retain their frozen ordering there; on RoboCasa, DINOv3 overtakes DINOv2 and VC-1. A strong starting point and a large return to adaptation are distinct advantages (Appendix~\ref{sec:results-ft}).

\subsection{Multi-task performance across encoders}

\begin{table}[t]
\caption{\textbf{Frozen multi-task policies.} Twelve-task mean success with propagated evaluation deviations; Rk is rank. \best{First}, \secondbest{second}, and \thirdbest{third} are highlighted. $N\times D$: tokens per view and context width; $^{\dagger}$VLM features are projected to 1,024 channels. Training cost is per RoboCasa MT step. Family colors follow Table~\ref{tab:encoder_families}.}
\label{tab:suite}
\centering\begingroup\footnotesize
\setlength{\tabcolsep}{2.1pt}\renewcommand{\arraystretch}{1.05}
\begin{tabular*}{\linewidth}{@{\extracolsep{\fill}}lrr cc cc rr@{}}
\toprule
& & & \multicolumn{2}{c}{\textbf{RoboCasa MT}} & \multicolumn{2}{c}{\textbf{RoboTwin MT}} & \multicolumn{2}{c}{\textbf{Training / step}}\\
\cmidrule(lr){4-5}\cmidrule(lr){6-7}\cmidrule(lr){8-9}
\textbf{Encoder} & \textbf{Params} & $\boldsymbol{N\times D}$ & \textbf{SR} & \textbf{Rk} & \textbf{SR} & \textbf{Rk} & \textbf{TFLOPs} & \textbf{s}\\\midrule
\textcolor{hiveTeal}{VGGT-$\Omega$} & 1B & $196\times 2,048$ & \best{35.39{\scriptsize\,$\pm$1.33}} & 1 & 50.00{\scriptsize\,$\pm$1.33} & 4 & 113.3 & 0.869\\
\textcolor{hiveTeal}{LingBot-Vision} & 300M & $196\times 1,024$ & 29.39{\scriptsize\,$\pm$1.49} & 5 & 46.78{\scriptsize\,$\pm$1.37} & 6 & 65.5 & 0.676\\
\textcolor{hiveTeal}{SPA} & 86M & $196\times 768$ & 16.50{\scriptsize\,$\pm$1.14} & 17 & 12.33{\scriptsize\,$\pm$0.95} & 20 & 49.7 & 0.565\\
\midrule[0.25pt]
\textcolor{hiveAccent}{DINOv2} & 86M & $256\times 768$ & \thirdbest{32.28{\scriptsize\,$\pm$1.45}} & 3 & 49.39{\scriptsize\,$\pm$1.04} & 5 & 54.8 & 0.662\\
\textcolor{hiveAccent}{DINOv3} & 86M & $196\times 768$ & 31.22{\scriptsize\,$\pm$1.34} & 4 & 30.44{\scriptsize\,$\pm$1.44} & 13 & 49.8 & 0.597\\
\textcolor{hiveAccent}{MAE} & 86M & $196\times 768$ & 25.28{\scriptsize\,$\pm$1.49} & 8 & 26.67{\scriptsize\,$\pm$1.56} & 15 & 49.7 & 0.576\\
\textcolor{hiveAccent}{V-JEPA~2.1} & 80M & $576\times 768$ & 18.67{\scriptsize\,$\pm$1.22} & 15 & 39.61{\scriptsize\,$\pm$1.28} & 10 & 83.0 & 0.923\\
\midrule[0.25pt]
\textcolor{hiveRose}{Xiaomi-Robotics-1} & 4.4B & $64\times 1,024$$^{\dagger}$ & 25.17{\scriptsize\,$\pm$1.70} & 9 & \secondbest{53.22{\scriptsize\,$\pm$1.30}} & 2 & 195.3 & 1.120\\
\textcolor{hiveRose}{DepthVLM} & 4B & $64\times 1,024$$^{\dagger}$ & 18.50{\scriptsize\,$\pm$1.08} & 16 & \best{53.67{\scriptsize\,$\pm$1.24}} & 1 & 195.3 & 1.138\\
\textcolor{hiveRose}{Qwen3-VL} & 4B & $64\times 1,024$$^{\dagger}$ & 19.56{\scriptsize\,$\pm$1.19} & 14 & \thirdbest{51.44{\scriptsize\,$\pm$1.17}} & 3 & 195.3 & 1.132\\
\midrule[0.25pt]
\textcolor{hiveAmber}{SigLIP2} & 86M & $196\times 768$ & 29.11{\scriptsize\,$\pm$1.53} & 6 & 44.22{\scriptsize\,$\pm$1.24} & 8 & 49.7 & 0.581\\
\textcolor{hiveAmber}{InternViT} & 300M & $256\times 1,024$ & 20.11{\scriptsize\,$\pm$0.97} & 13 & 30.94{\scriptsize\,$\pm$1.28} & 12 & 75.1 & 0.663\\
\textcolor{hiveAmber}{SigLIP} & 86M & $196\times 768$ & 13.67{\scriptsize\,$\pm$1.28} & 18.5 & 28.39{\scriptsize\,$\pm$1.37} & 14 & 49.7 & 0.595\\
\textcolor{hiveAmber}{CLIP} & 86M & $196\times 768$ & 11.44{\scriptsize\,$\pm$0.98} & 20 & 21.33{\scriptsize\,$\pm$0.84} & 18 & 49.7 & 0.589\\
\midrule[0.25pt]
\textcolor{hiveCyan}{VC-1} & 86M & $196\times 768$ & 28.44{\scriptsize\,$\pm$1.42} & 7 & 25.06{\scriptsize\,$\pm$1.46} & 16 & 49.7 & 0.570\\
\textcolor{hiveCyan}{Voltron} & 86M & $196\times 768$ & 24.83{\scriptsize\,$\pm$1.01} & 10.5 & 22.72{\scriptsize\,$\pm$1.04} & 17 & 51.9 & 0.602\\
\midrule[0.25pt]
\textcolor{hiveIndigo}{Theia} & 86M & $196\times 768$ & 24.83{\scriptsize\,$\pm$1.08} & 10.5 & 43.83{\scriptsize\,$\pm$0.87} & 9 & 49.7 & 0.610\\
\textcolor{hiveIndigo}{C-RADIOv4} & 631M & $196\times 1,280$ & \secondbest{34.78{\scriptsize\,$\pm$1.43}} & 2 & 45.17{\scriptsize\,$\pm$1.39} & 7 & 87.8 & 0.669\\
\textcolor{hiveIndigo}{RADIOv2.5} & 98M & $196\times 768$ & 24.39{\scriptsize\,$\pm$1.11} & 12 & 31.11{\scriptsize\,$\pm$1.19} & 11 & 49.8 & 0.565\\
\midrule[0.25pt]
\textcolor{hiveMuted}{ViT} & 86M & $196\times 768$ & 13.67{\scriptsize\,$\pm$1.23} & 18.5 & 18.61{\scriptsize\,$\pm$0.99} & 19 & 49.7 & 0.577\\
\bottomrule\end{tabular*}\endgroup
\end{table}

\paragraph{\textcolor{hiveAccent}{Strong encoders emerge from several pretraining families.}}\label{fnd:bands}
We next hold the patch-conditioned interface fixed and compare the full encoder suite under MT (Table~\ref{tab:suite}). VGGT-$\Omega$ leads RoboCasa at 35.39\%, followed by C-RADIOv4 at 34.78\% and DINOv2 at 32.28\%. The three VLMs lead RoboTwin at 51.44--53.67\%. Cross-benchmark ranks agree only moderately ($\rho=0.43$): VC-1 outperforms V-JEPA~2.1 on RoboCasa but trails it on RoboTwin. Pretraining family alone is too broad a label to settle the choice; the target setting changes which strengths matter.

Cost also changes the choice. DINOv2 comes within 3.11 points of VGGT-$\Omega$ on RoboCasa at 54.8 versus 113.3 TFLOPs per step, while the VLMs lead RoboTwin at 195.3 TFLOPs per step. A shared optimizer schedule therefore does not make the encoders equally expensive to train, and the cheaper run can be the practical choice when the success gap is only a few points.

\subsection{Sharing a policy makes broad task coverage affordable}
\label{sec:m-sharing}

\paragraph{\textcolor{hiveAccent}{One shared policy preserves most of the success of task-specific policies.}}
To measure the price of sharing, we compare RoboCasa ST and MT with identical demonstrations per task and 100k updates per policy (Figure~\ref{fig:m-closedloop}c). The average falls from 26.94\% to 23.62\%, retaining 87.7\% of ST success. Losses vary: 0.67 points for SigLIP2, 1.28 for DINOv3, and 7.78 for VGGT-$\Omega$; MAE and Xiaomi-Robotics-1 improve slightly.

One MT run covers all twelve tasks with one twelfth of the optimizer steps. Each task receives fewer updates, yet most of its success survives (Appendix~\ref{sec:results-sharing}). The saving is in updates per suite, not in demonstrations per task, which stay matched between the two regimes.

\subsection{More demonstrations help more consistently than more parameters}
\label{sec:m-scale}

\begin{figure}[t]
\centering
\includegraphics[width=\linewidth]{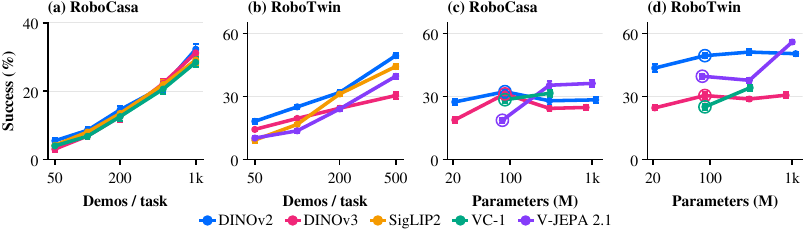}
\caption{\textbf{Scaling multi-task policies.} Demonstrations per task (a, b) and encoder size (c, d); rings mark default checkpoints and error bars propagate evaluation deviations. Each policy uses the same update budget. The four-task RoboCasa ST sweep is in Appendix~\ref{sec:results-data}.}
\label{fig:m-scale}
\end{figure}

\paragraph{\textcolor{hiveAccent}{Additional demonstrations improve every curve.}}
We retrain policies on nested demonstration subsets with the same update budget. Increasing demonstrations per task from 50 to 1,000 adds 24.44--28.16 points across RoboCasa MT encoders. All RoboTwin curves improve from 50 to 500; SigLIP2 gains 34.94 points and overtakes DINOv3 (Figure~\ref{fig:m-scale}a,b). On matched RoboCasa tasks C1--C4, doubling 500 to 1,000 adds 10.33--13.00 points for MT, versus $-0.17$ to $+3.83$ for ST. Shared policies still benefit where task-specific curves flatten (Appendix~\ref{sec:results-data}).

\paragraph{\textcolor{hiveAccent}{The value of model scale depends on the family.}}
Within-family sweeps show a less uniform return. DINOv2 improves from Small to Base, then plateaus on RoboTwin and declines on RoboCasa; DINOv3 Base is its family's strongest RoboCasa checkpoint. V-JEPA~2.1 instead rises from 18.67\% to 36.22\% on RoboCasa between Base and giant, and reaches 55.89\% on RoboTwin. VC-1 Large improves on both suites (Figure~\ref{fig:m-scale}c,d). Scale helps when the additional capacity transfers to the policy.

\subsection{Input and extraction choices have different returns}
\label{sec:m-inputs}

\begin{figure}[!b]\centering
\includegraphics[width=\linewidth]{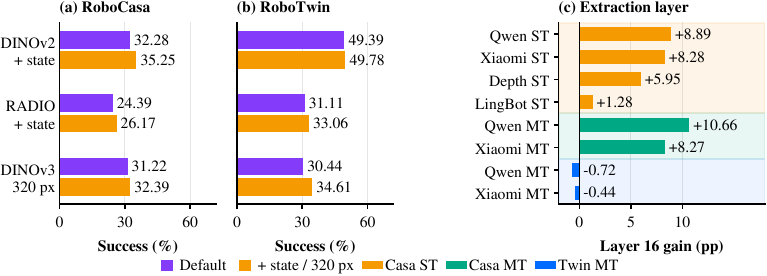}
\caption{\textbf{Changing inputs and the extraction layer.} (a, b) MT success with robot state or DINOv3 at 320 pixels (400 versus 196 tokens per view). (c) Layer-16 minus final-layer success; ST and MT denote the policy regime. Layer choice has larger effects on RoboCasa than the input additions.}
\label{fig:state-resolution}
\end{figure}

\paragraph{\textcolor{hiveAccent}{Robot state and finer image grids give smaller gains.}}
Two controls test whether missing proprioception or a coarser grid explains the encoder gaps. A state token improves DINOv2 and RADIOv2.5 by 0.39--2.97 points and preserves their ordering (Figure~\ref{fig:state-resolution}a,b). More than doubling DINOv3's tokens, from 196 to 400, adds 1.17 and 4.17 points. Both effects are smaller than retaining patches in the first place, so the image stream, not a missing state token, carries most of the gap.

\paragraph{\textcolor{hiveAccent}{An intermediate layer can recover much of a VLM's RoboCasa deficit.}}
Replacing final-layer features with layer 16 tests whether useful information is easier to read earlier in the network. Qwen3-VL and Xiaomi gain 8--11 points on RoboCasa and change by less than a point on RoboTwin (Figure~\ref{fig:state-resolution}c). The useful layer is not the same for every model, just as the useful size is not the same for every family (Appendices~\ref{sec:results-register}--\ref{sec:results-layer}).

\section{What Tracks Closed-Loop Success?}
\label{sec:m-diag}

\begin{figure}[!b]\centering
\includegraphics[width=\linewidth]{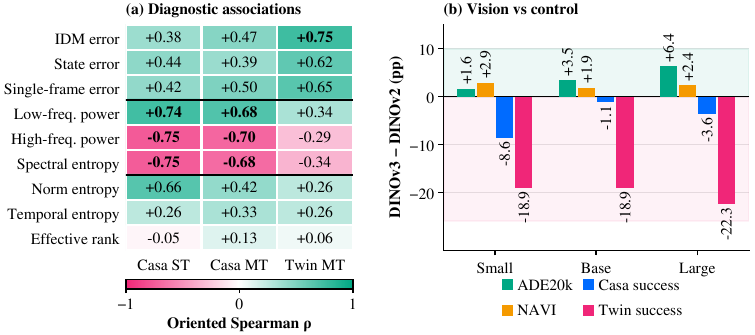}
\caption{\textbf{Different settings favor different diagnostics.} (a) Associations with success among vision encoders. Error signs are reversed; other metrics retain their signs. Bold cells have $q<0.05$ among primary diagnostics. (b) DINOv3 minus DINOv2 at matched sizes: vision gains and control losses point in opposite directions. Differences use each metric's score points.}
\label{fig:m-landscape}\end{figure}

Prior work connects an encoder's value for control to its ability to recover scene structure, robot state, or actions~\citep{burns,spa,structure,semanticwm}. We compare these readouts with label-free token statistics and published vision scores, then ask which relationships persist across settings and adaptation. Figure~\ref{fig:m-landscape} summarizes the main contrasts; Appendix~\ref{app:diagnostics} reports all associations.

\subsection{Action readout is informative, but shares a static component}
\label{sec:m-readout}

\begin{table}[!htb]\centering\footnotesize
\caption{\textbf{Readout error versus success.} Raw Spearman $\rho$; darker green marks stronger negative associations. Bold entries have $q<0.05$ among primary diagnostics. Adding VLMs preserves the state association but largely removes the object association. Cohorts and single-frame control tests are detailed in Appendix~\ref{app:stats}.}
\label{tab:m-robust}\setlength{\tabcolsep}{4pt}
\begin{tabular}{@{}p{.28\linewidth}*{6}{>{\centering\arraybackslash}p{.10\linewidth}}@{}}\toprule
\rowcolor{hiveAccent!9}&\multicolumn{3}{c}{Vision encoders}&\multicolumn{3}{c}{Including VLMs}\\
\cmidrule(lr){2-4}\cmidrule(lr){5-7}
Readout & Casa ST & Casa MT & Twin MT & Casa ST & Casa MT & Twin MT\\\midrule
Two-frame IDM & \cellcolor{hiveTeal!21}-0.38 & \cellcolor{hiveTeal!24}-0.47 & \cellcolor{hiveTeal!34}\textbf{-0.75} & \cellcolor{hiveTeal!18}-0.28 & \cellcolor{hiveTeal!27}-0.53 & \cellcolor{hiveTeal!33}\textbf{-0.70}\\
Robot state & \cellcolor{hiveTeal!23}-0.44 & \cellcolor{hiveTeal!22}-0.39 & \cellcolor{hiveTeal!30}-0.62 & \cellcolor{hiveTeal!27}-0.53 & \cellcolor{hiveTeal!26}-0.50 & \cellcolor{hiveTeal!25}-0.50\\
Single-frame IDM & \cellcolor{hiveTeal!23}-0.42 & \cellcolor{hiveTeal!26}-0.50 & \cellcolor{hiveTeal!31}-0.65 & \cellcolor{hiveTeal!20}-0.36 & \cellcolor{hiveTeal!26}-0.53 & \cellcolor{hiveTeal!28}-0.56\\
Object position & \cellcolor{black!4}-- & \cellcolor{black!4}-- & \cellcolor{hiveTeal!27}-0.54 & \cellcolor{black!4}-- & \cellcolor{black!4}-- & \cellcolor{hiveTeal!10}-0.05\\
\bottomrule\end{tabular}\end{table}

\paragraph{\textcolor{hiveTeal}{IDM is one of the strongest readouts of closed-loop success.}}\label{fnd:idm}
We ask which cheap measurement best ranks frozen encoders before a policy is trained. Among vision encoders, two-frame IDM error correlates with RoboTwin MT success at $\rho=-0.75$ (95\% interval $[-0.90,-0.40]$, $q=0.020$), and removing any pretraining family leaves $\rho$ between $-0.88$ and $-0.70$. The association keeps the expected sign on the tasks we test. IDM is thus a strong correlate on RoboTwin, although even this association does not reliably identify the best encoder for control.

\paragraph{\textcolor{hiveTeal}{Matching probe capacity preserves the state association.}}\label{fnd:idm-static}
With the same width-256 transformer, state error correlates with success at $-0.44$, $-0.39$, and $-0.62$ across the three vision cohorts. Including VLMs leaves associations of $-0.53$, $-0.50$, and $-0.50$ (Table~\ref{tab:m-robust}). Object error behaves differently: its RoboTwin correlation changes from $-0.54$ to $-0.05$ with VLMs. Recovering robot configuration and object position therefore give different assessments of the same representations.

\paragraph{\textcolor{hiveTeal}{A future frame improves prediction, but does not isolate motion.}}
Because the target includes $s_t$, IDM and state errors move together ($\rho=0.80$ and $0.82$). A future frame lowers error for every frozen encoder, by 40.7\% and 46.9\% at the median, but the encoder order barely changes. The second view therefore adds information without changing which encoder looks best. This control tests access to a later observation, rather than isolating motion understanding.

\begin{figure}[!htb]\centering
\includegraphics[width=\linewidth]{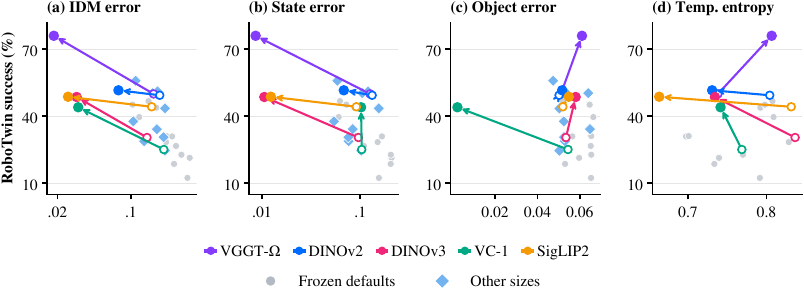}
\caption{\textbf{Following encoders through fine-tuning on RoboTwin.} Arrows join frozen (open) and adapted (filled) checkpoints. Gray points show frozen defaults; blue diamonds add other model sizes. Action error falls consistently, while object error and temporal entropy need not improve with success. The first two horizontal axes are logarithmic.}
\label{fig:m-shift}\end{figure}

\paragraph{\textcolor{hiveTeal}{Fine-tuning improves action readout without uniformly improving scene readout.}}\label{fnd:adapt}
Repeating the readouts after policy learning separates adaptation from cross-encoder differences. All five paired RoboTwin encoders improve success and lower both IDM and state error, yet four increase object error (Figure~\ref{fig:m-shift}). The gain in success and the gain in object readout are not the same change. Policy learning favors action-relevant information without improving every scene readout.

\suppressfloats[t]

Pooling adapted checkpoints with frozen encoders strengthens the IDM correlation to $-0.80$, versus $-0.40$ within the adapted group. Frozen IDM error does not order adapted success. The pooled relationship mostly separates frozen from adapted representations, so it is a weak guide to which encoder to fine-tune. A strong correlation across the pooled set can be read as evidence about adaptation when it is only evidence about the frozen starting points.

\paragraph{\textcolor{hiveTeal}{Broad associations also leave room for size and layer reversals.}}
Across size variants the IDM--success correlation remains negative, yet DINOv2 Base to giant lowers IDM error while success falls. Qwen and Xiaomi likewise improve both readouts at layer 16 on RoboTwin, with little change in success. Size and layer are changes inside one model, and a cross-encoder correlation need not survive them. A readout can separate pretrained encoders without predicting every change inside one model.

\subsection{Better vision scores do not reliably imply better control}
\label{sec:m-vision}\label{fnd:vision}

Published vision scores provide no consistent ordering for control. Matched checkpoints for classification, segmentation, depth, correspondence, and label propagation~\citep{lingbot,dinov3,radio} show no nominal association with multi-task success on either suite (Appendix~\ref{app:vision}). ImageNet linear-probe accuracy gives $\rho=0.07$ on RoboCasa and $0.38$ on RoboTwin.

Matched sizes sharpen the contrast. DINOv3 gains 1.65--6.37 mIoU on ADE20k and 1.9--2.9 NAVI points over DINOv2, yet loses 1.06--8.56 success points on RoboCasa and 18.94--22.33 on RoboTwin (Figure~\ref{fig:m-landscape}b). A high score on a vision benchmark can sit next to a strong policy, but the improvement on the benchmark is not the improvement in control.

\subsection{Token structure is informative within a setting}
\label{sec:m-tokens}\label{fnd:tokens}

\paragraph{\textcolor{hiveTeal}{Low-frequency spatial structure stands out on RoboCasa.}}
Low-frequency spatial power correlates with success at $+0.74$ for ST and $+0.68$ for MT, and the complementary high-frequency measures survive correction on both RoboCasa regimes. These summaries describe one spectrum, not four causes. On RoboTwin the low-frequency association falls to $+0.34$ and does not survive correction. The association describes the tokens the policy reads; it does not say that smoothing a feature map would help.

\paragraph{\textcolor{hiveTeal}{Norm entropy is useful mainly before adaptation.}}
Norm-histogram entropy correlates with frozen success at $+0.66$, $+0.42$, and $+0.26$, then temporal norm entropy falls in four of five fine-tuned pairs even as success rises. Covariance rank is similar: the participation ratio is essentially uncorrelated with frozen success, and fine-tuning collapses a 3.15--23.62 range into 5.67--7.04. Neither a diverse norm histogram nor a high-dimensional feature cloud is a universal control objective.

\subsection{Combining diagnostics and selecting candidates}
\label{sec:m-combine}\label{fnd:combo}

\begin{table}[t]\centering\footnotesize
\caption{\textbf{Predicting a held-out encoder family.} Nested ridge regression; MAE is in success points, and $\downarrow$\% is error reduction relative to the mean baseline. Green shading shows the reduction; bold marks the lowest MAE. ``All four'' adds temporal norm entropy to IDM, state, and norm entropy.}
\label{tab:diag-predict}\setlength{\tabcolsep}{4pt}
\begin{tabular}{@{}p{.28\linewidth}*{6}{>{\centering\arraybackslash}p{.10\linewidth}}@{}}\toprule
\rowcolor{hiveAccent!9}&\multicolumn{2}{c}{Casa ST}&\multicolumn{2}{c}{Casa MT}&\multicolumn{2}{c}{Twin MT}\\
\cmidrule(lr){2-3}\cmidrule(lr){4-5}\cmidrule(lr){6-7}
Predictor & MAE & $\downarrow$\% & MAE & $\downarrow$\% & MAE & $\downarrow$\%\\\midrule
Mean baseline & \cellcolor{black!4}8.24 & -- & \cellcolor{black!4}7.03 & -- & \cellcolor{black!4}10.91 & --\\
IDM & \cellcolor{black!4}8.25 & -0.2 & \cellcolor{hiveTeal!10}6.69 & +4.9 & \cellcolor{hiveTeal!34}\textbf{6.73} & +38.3\\
State & \cellcolor{hiveTeal!11}7.76 & +5.8 & \cellcolor{hiveTeal!12}6.56 & +6.7 & \cellcolor{hiveTeal!32}7.00 & +35.9\\
IDM + State & \cellcolor{hiveTeal!11}7.82 & +5.1 & \cellcolor{hiveTeal!18}5.94 & +15.6 & \cellcolor{hiveTeal!26}7.95 & +27.1\\
IDM + entropy & \cellcolor{hiveTeal!15}\textbf{7.25} & +12.0 & \cellcolor{hiveTeal!19}\textbf{5.86} & +16.7 & \cellcolor{hiveTeal!29}7.50 & +31.2\\
All four & \cellcolor{hiveTeal!11}7.75 & +5.9 & \cellcolor{black!4}7.06 & -0.4 & \cellcolor{hiveTeal!26}8.01 & +26.6\\
\bottomrule\end{tabular}\end{table}

\paragraph{\textcolor{hiveTeal}{More diagnostics help selectively.}}
Nested family-held-out prediction tests whether diagnostics complement one another (Table~\ref{tab:diag-predict}). IDM lowers RoboTwin MAE from 10.91 to 6.73 points, and state or entropy adds no further benefit there. Adding every measurement at once loses to the best smaller subset in every setting. Extra measurements help only when their information transfers, so the gain comes from keeping the measurement that is new in that setting, not from collecting a longer list.

\paragraph{\textcolor{hiveTeal}{A useful screen need not identify the best encoder.}}
The five lowest-IDM encoders retain 53.22\% RoboTwin success, 0.45 points below the best, using one quarter of the runs. The same screen falls below random selection on both RoboCasa regimes (Table~\ref{tab:screening}). A shortlist can therefore sit close to the best encoder and still miss it, which is why the remaining candidates are checked with policy runs.

\paragraph{\textcolor{hiveTeal}{The benchmark matters more than task sharing.}}
Profiles agree much more closely within RoboCasa ($r=0.95$) than between the two benchmarks ($r=0.65$), so changing the robot and the views changes which properties accompany success. Sharing the tasks barely changes that profile, and a diagnostic chosen on single-task data remains informative when those tasks are trained together.

\section{Discussion and Conclusion}
\label{sec:m-disc}

HIVE-Bench connects dense egocentric representations to the policies that use them. Preserving patches, adding demonstrations, and adapting features improve control, and task sharing broadens coverage cheaply. IDM is one of the strongest guides to which frozen encoder will succeed, but no measurement is stable enough to prove that choice.

Three physical tasks preserve DINOv2's lead over DINOv3 and CLIP (Appendix~\ref{app:realworld}). The other evidence uses one head and two suites, so more tasks and seeds are still needed.

The open question is how to improve on IDM while keeping closed-loop success as the final measure.

\bibliographystyle{unsrtnat}
\bibliography{main}
\clearpage
\appendix
\section*{Appendix}
The appendix documents the benchmark and policy interfaces, provides the paired and ablation studies from the main text, and lists the task-level success rates. Training-seed variation and evaluation deviations are reported separately.

\section{Benchmark and Policy Details}
\label{sec:design}\label{app:protocol}\label{sec:benchmarks}\label{sec:encoders}

RoboCasa-GR1 uses a humanoid head camera and 29 action coordinates (two 7-DoF arms, two 6-DoF hands, and three waist coordinates). RoboTwin uses head and wrist cameras with 14 absolute joint targets (two 6-DoF arms and two grippers). Actions are min--max normalized. The loader stores views at $224\times224$ before encoder-specific preprocessing. Demonstration sweeps select complete episodes with a fixed subset seed; the smaller subsets are nested in the larger ones. Table~\ref{tab:app-tasks} lists all tasks.

\begin{table}[!ht]\centering\footnotesize
\caption{\textbf{Encoder families and primary references.} Colors group the default encoders in Table~\ref{tab:suite}.}
\label{tab:encoder_families}\setlength{\tabcolsep}{5pt}
\begin{tabular*}{\linewidth}{@{\extracolsep{\fill}}l>{\raggedright\arraybackslash}p{.76\linewidth}@{}}\toprule
Family & Encoders\\\midrule
\textcolor{hiveTeal}{Geometry} & VGGT-$\Omega$, LingBot-Vision, SPA~\citep{vggtomega,lingbot,spa}\\
\textcolor{hiveAccent}{Self-supervised} & DINOv2, DINOv3, MAE, V-JEPA~2.1~\citep{dinov2,dinov3,mae,vjepa21}\\
\textcolor{hiveRose}{VLM-derived} & Xiaomi-Robotics-1, DepthVLM, Qwen3-VL~\citep{xiaomi,depthvlm,qwen3}\\
\textcolor{hiveAmber}{Vision--language} & SigLIP2, InternViT, SigLIP, CLIP~\citep{siglip2,internvl,siglip,clip}\\
\textcolor{hiveCyan}{Robot-oriented} & VC-1, Voltron~\citep{cortex,voltron}\\
\textcolor{hiveIndigo}{Multi-teacher} & Theia, C-RADIOv4, RADIOv2.5~\citep{theia,radio,radiov25}\\
\textcolor{hiveMuted}{Supervised} & ViT~\citep{vit}\\\bottomrule
\end{tabular*}\end{table}

\subsection{Policy architecture}
\label{sec:policy}

Figure~\ref{fig:m-policy} shows the shared policy; the context interfaces and flow objective are specified below.

\paragraph{Context and attention.}
Context $C_t$ concatenates all views' patch tokens at native width $D_{\rm enc}$. In MT, task instructions are encoded as frozen CLIP ViT-B/32 text features, batch-padded and projected to $D_{\rm enc}$. The VLMs and LingBot-Vision project image features and the 77 text tokens of frozen CLIP ViT-B/16 to width 1,024. These interfaces retain text in ST, with a constant task instruction. Image tokens precede the instruction in the VLM chat template and do not encode it.

The DiT has width 768, 16 blocks, 12 attention heads, and dropout 0.2. Its input sequence $H_t$ contains 32 learned queries and 16 embedded action tokens with learned positions. Eight blocks cross-attend to visual/text context, alternating with eight self-attention blocks:
\begin{equation}
\mathrm{CrossAttn}(H_t,C_t)=\mathrm{softmax}\!\left((H_tW_Q)(C_tW_K)^\top/\sqrt{64}\right)C_tW_V.
\label{eq:cross_attention}
\end{equation}
Parameter counts, in millions for the action head and all downstream weights, are 138.669/139.063 for thirteen 768-width encoders, 142.231/142.756 for InternViT, and 145.793/146.449 for C-RADIOv4. VGGT-$\Omega$ uses 156.478/156.478. Qwen3-VL, Xiaomi, and DepthVLM use 142.690/145.838, and LingBot-Vision uses 142.690/144.265. Counts exclude pretrained encoders but include the allocated state MLP even when unused; downstream adds external projections. Thirteen heads match exactly; the largest has 12.8\% more parameters. Compute varies (Table~\ref{tab:suite}).

\paragraph{Flow matching.}
For normalized action chunk $A_t$, draw $\epsilon\sim\mathcal N(0,I)$ and $b\sim\mathrm{Beta}(1.5,1)$, set $\tau=(s-\min(b,s))/s$ with $s=0.999$, and interpolate $A_\tau=(1-\tau)\epsilon+\tau A_t$. The objective is
\begin{equation}
\mathcal L_{\rm act}=\mathbb E\left[\|v_\theta(A_\tau,\lfloor1000\tau\rfloor,C_t)-(A_t-\epsilon)\|_F^2/(16d_a)\right].
\label{eq:policy_loss}
\end{equation}
Each example receives eight noise/time draws per update. Inference starts from Gaussian noise and integrates $v_\theta$ for 15 Euler steps on RoboCasa or 10 on RoboTwin. RoboCasa executes the first 12 predicted actions before replanning, with a maximum of 720 control steps per episode. RoboTwin executes all 16, without temporal ensembling, in randomized scenes with unseen instructions.

\subsection{Optimization and evaluation}
\label{sec:settings}

\begin{table}[!ht]\centering\small
\caption{\textbf{Training configuration.} Encoder learning rate applies during fine-tuning. Other trainable parameters include the context projections.}
\label{tab:defaults}\setlength{\tabcolsep}{7pt}
\begin{tabular*}{\linewidth}{@{\extracolsep{\fill}}lccc@{}}\toprule
Setting & RoboCasa ST & RoboCasa MT & RoboTwin MT\\\midrule
Tasks per policy & 1 & 12 & 12\\
Demonstrations per task & 1,000 & 1,000 & 500\\
Optimizer steps & 100k & 100k & 100k\\
Global batch / GPUs / per-GPU batch & 128 / 4 / 32 & 128 / 4 / 32 & 128 / 4 / 32\\
Gradient accumulation & None & None & None\\
Action head learning rate & $10^{-4}$ & $10^{-4}$ & $10^{-4}$\\
Encoder learning rate (FT) & $10^{-5}$ & $10^{-5}$ & $10^{-5}$\\
Other trainable parameters LR & $10^{-4}$ & $3\times10^{-5}$ & $10^{-4}$\\
Warmup steps & 3,000 & 3,000 & 5,000\\
Learning-rate floor & $5\times10^{-8}$ & $5\times10^{-8}$ & $5\times10^{-7}$\\
Scheduler & \multicolumn{3}{c}{Linear warmup + cosine decay}\\
Optimizer & \multicolumn{3}{c}{AdamW, $\beta=(0.9,0.95)$, $\epsilon=10^{-8}$}\\
Weight decay / clipping / dropout & \multicolumn{3}{c}{$10^{-8}$ / 1.0 / 0.2}\\
Precision & BF16 & BF16 & BF16\\\bottomrule
\end{tabular*}\end{table}

Training uses gradient checkpointing and DeepSpeed ZeRO stage 2. Frozen runs update the action head and projections; FT also updates the visual encoder at one tenth of the head's learning rate. Text encoders remain frozen. Covering twelve tasks uses twelve ST policies or one MT policy, so equal 100k-step budgets per policy correspond to 1.2M versus 100k total updates.

The final checkpoint is evaluated with three evaluation seeds and 50 rollouts per task per seed. Task success is the mean across seeds, with the exported evaluation standard deviation $\sigma_k$. Suite success averages the $K$ task scores. Main and scale tables propagate deviations as $\sqrt{\sum_k\sigma_k^2}/K$, with $K=12$ or $4$ for the ST data sweep. This assumes independent task errors and describes evaluation variability. State, resolution, and query ablations retain their supplied suite deviations, which are not all task-propagated. The independent three-training-seed experiment reports a population SD across training runs (Table~\ref{tab:train-seeds}).

\section{Encoder Inventory and Extraction Interfaces}
\label{app:encoders}

Tables~\ref{tab:pretrain} and~\ref{tab:interfaces} document the pretraining and extraction interface of each default checkpoint. Table~\ref{tab:scale-variants} lists the model-scale variants. Each encoder receives its prescribed image normalization once, and the action head reads the resulting spatial tokens.

\begin{table}[!ht]
\centering
\caption{\textbf{Pretraining data, amount, and objective.} ``Native'' is the resolution at which the released checkpoint was trained; ``multi-res.'' marks training over several resolutions. Encoder names are colored by family as in Table~\ref{tab:suite}.}
\label{tab:pretrain}\nocitehere
\begingroup
\arrayrulecolor{hiveInk}
\setlength{\tabcolsep}{3.2pt}
\renewcommand{\arraystretch}{1.12}
\scriptsize
\begin{tabular*}{\linewidth}{@{\extracolsep{\fill}}>{\raggedright\arraybackslash}p{0.13\linewidth}>{\raggedright\arraybackslash}p{0.41\linewidth}>{\raggedright\arraybackslash}p{0.33\linewidth}>{\raggedright\arraybackslash}p{0.07\linewidth}@{}}
\toprule[0.8pt]
\textbf{Encoder} & \textbf{Pretraining data and amount} & \textbf{Objective} & \textbf{Native}\\
\midrule[0.45pt]
\textcolor{hiveTeal}{VGGT-$\Omega$}~\cite{vggtomega} & About 4M labeled multi-view sequences (about 3M from 30+ public and internal datasets, 0.8M annotated Internet videos) and 18M unlabeled videos & Camera, depth, and point-map prediction with one dense head, then teacher--student self-distillation on unlabeled video; DINOv3-initialized ViT & 512\\
\textcolor{hiveTeal}{LingBot-Vision}~\cite{lingbot} & About 161M images curated from a pool of about 2B web images; the ViT-L is distilled from a 1.1B ViT-g/16 teacher & Masked boundary modeling on DINO/iBOT-style self-distillation, with Gram anchoring and high-resolution adaptation & Multi-res.\\
\textcolor{hiveTeal}{SPA}~\cite{spa} & 3.8M frames from ScanNet, ScanNet++, Hypersim, ADT, S3DIS, and DROID & Masked (50\%) multi-view volume rendering of RGB, depth, and semantics; initialized from ImageNet MAE & 224\\
\textcolor{hiveAccent}{DINOv2}~\cite{dinov2} & LVD-142M, 142M curated images; Base distilled from ViT-g/14 & DINO and iBOT self-distillation with KoLeo regularization & 224, 518\\
\textcolor{hiveAccent}{DINOv3}~\cite{dinov3} & LVD-1689M, about 1.7B curated images; Base distilled from a 7B teacher & DINO and iBOT self-distillation; teacher trained with Gram anchoring and high-resolution adaptation & Multi-res.\\
\textcolor{hiveAccent}{MAE}~\cite{mae} & ImageNet-1K, 1.28M images & Masked (75\%) pixel reconstruction & 224\\
\textcolor{hiveAccent}{V-JEPA~2.1}~\cite{vjepa21} & VisionMix-163M: LVD-142M images with SSv2 (168K), Kinetics (733K), HowTo100M (1.1M), and YT-1B (19M) videos; Base distilled from a 2B ViT-G & Masked latent prediction with a dense loss on visible and masked tokens and deep self-supervision & 384\\
\textcolor{hiveRose}{Xiaomi-Robotics-1}~\cite{xiaomi} & Qwen3-VL initialization; over 100K hours of real-world manipulation trajectories for pretraining and over 10K hours of cross-embodiment data for post-training & Vision--language--action training of a mixture-of-transformers VLM with a flow-matching action DiT & Dynamic\\
\textcolor{hiveRose}{DepthVLM}~\cite{depthvlm} & Qwen3-VL-4B initialization; 4.4M samples from eight metric-depth datasets (ScanNet++, Taskonomy, HM3D, Matterport3D, Argoverse2, Waymo, DDAD, nuScenes) & Language and dense metric-depth losses through a 34M DPT head; head-only, then end-to-end training & Dynamic\\
\textcolor{hiveRose}{Qwen3-VL}~\cite{qwen3} & About 2.2T multimodal pretraining tokens (67B alignment, two stages of about 1T, 100B long context), then instruction post-training & Multimodal next-token prediction; SigLIP2-Large vision encoder with DeepStack injection into the LLM & Dynamic\\
\textcolor{hiveAmber}{SigLIP2}~\cite{siglip2} & WebLI, 10B images with 12B alt-texts in 109 languages & Sigmoid image--text loss with a captioning decoder, self-distillation, and masked prediction & 224\\
\textcolor{hiveAmber}{InternViT}~\cite{internvl} & Distilled from InternViT-6B, then trained inside InternVL; the V2.5 release adds incremental training on OCR, chart, math, and document data & Cosine feature distillation, then next-token prediction through an LLM & 448, dynamic\\
\textcolor{hiveAmber}{SigLIP}~\cite{siglip} & English image--text pairs of WebLI & Sigmoid image--text loss & 224\\
\textcolor{hiveAmber}{CLIP}~\cite{clip} & WIT, 400M image--text pairs & Contrastive image--text loss & 224\\
\textcolor{hiveCyan}{VC-1}~\cite{cortex} & Ego4D, manipulation and navigation video, and ImageNet; 5.6M frames & Masked autoencoding & 224\\
\textcolor{hiveCyan}{Voltron}~\cite{voltron} & Something-Something-v2 with captions, about 168K clips at five frames per clip and epoch & Language-conditioned masked autoencoding of single frames (V-Cond) & 224\\
\textcolor{hiveIndigo}{Theia}~\cite{theia} & ImageNet-1K, 1.28M images; trained from scratch & Feature distillation from CLIP, DINOv2, and ViT through shallow CNN translators & 224\\
\textcolor{hiveIndigo}{C-RADIOv4}~\cite{radio} & About 700M images with commercially permissive licenses & Multi-teacher distillation from SigLIP2-g, DINOv3-7B, and SAM3 & Multi-res.\\
\textcolor{hiveIndigo}{RADIOv2.5}~\cite{radiov25} & DataComp-1B images & Multi-teacher distillation from DFN CLIP, SigLIP, DINOv2-g, and SAM-H & Multi-res.\\
\textcolor{hiveMuted}{ViT}~\cite{vit} & ImageNet-21K (14M images, 21,843 classes), then ImageNet-1K & Supervised classification & 224\\
\bottomrule[0.8pt]
\end{tabular*}
\endgroup
\end{table}

\begin{table}[!ht]
\centering
\caption{\textbf{Checkpoints and token interfaces of the default encoders.} Input is the resolution at which the encoder sees each view, $N$ the number of patch tokens per view, and $D$ the token width. All data are stored at $224\times224$; the preprocessing that produces each input is described in the text.}
\label{tab:interfaces}
\begingroup
\arrayrulecolor{hiveInk}
\setlength{\tabcolsep}{3pt}
\renewcommand{\arraystretch}{1.1}
\scriptsize
\begin{tabular*}{\linewidth}{@{\extracolsep{\fill}}l>{\raggedright\arraybackslash}p{0.36\linewidth}lrccr@{}}
\toprule[0.8pt]
\textbf{Encoder} & \textbf{Checkpoint} & \textbf{Backbone} & \textbf{Params} & \textbf{Input} & \textbf{$N$} & \textbf{$D$}\\
\midrule[0.45pt]
\textcolor{hiveTeal}{VGGT-$\Omega$} & \texttt{facebook/}\allowbreak\texttt{VGGT-Omega} (\texttt{vggt\_omega\_1b\_512}) & Aggregator, patch 16 & 1B & $224^2$ & 196 & 2{,}048$^{a}$\\
\textcolor{hiveTeal}{LingBot-Vision} & \texttt{robbyant/}\allowbreak\texttt{lingbot-vision-vit-large} & ViT-L/16 & 300M & $224^2$ & 196 & 1{,}024$^{b}$\\
\textcolor{hiveTeal}{SPA} & \texttt{HaoyiZhu/}\allowbreak\texttt{SPA} (\texttt{spa\_vit\_base\_patch16}) & ViT-B/16 & 86M & $224^2$ & 196 & 768\\
\textcolor{hiveAccent}{DINOv2} & \texttt{facebook/}\allowbreak\texttt{dinov2-base} & ViT-B/14 & 86M & $224^2$ crop & 256 & 768\\
\textcolor{hiveAccent}{DINOv3} & \texttt{facebook/}\allowbreak\texttt{dinov3-vitb16-pretrain-lvd1689m} & ViT-B/16, 4 registers & 86M & $224^2$ & 196 & 768\\
\textcolor{hiveAccent}{MAE} & \texttt{facebook/}\allowbreak\texttt{vit-mae-base} & ViT-B/16 & 86M & $224^2$ & 196 & 768\\
\textcolor{hiveAccent}{V-JEPA~2.1} & \texttt{vjepa2\_1\_vit\_base\_384} (torch hub) & ViT-B/16 & 80M & $384^2$ & 576 & 768\\
\textcolor{hiveRose}{Xiaomi-Robotics-1} & \texttt{XiaomiRobotics/}\allowbreak\texttt{Xiaomi-Robotics-1-5B}, VLM part & Qwen3-VL & 4.4B & $256^2$ & 64 & 1{,}024$^{b}$\\
\textcolor{hiveRose}{DepthVLM} & \texttt{JonnyYu828/}\allowbreak\texttt{DepthVLM-4B} & Qwen3-VL & 4B & $256^2$ & 64 & 1{,}024$^{b}$\\
\textcolor{hiveRose}{Qwen3-VL} & \texttt{Qwen/}\allowbreak\texttt{Qwen3-VL-4B-Instruct} & Qwen3-VL & 4B & $256^2$ & 64 & 1{,}024$^{b}$\\
\textcolor{hiveAmber}{SigLIP2} & \texttt{google/}\allowbreak\texttt{siglip2-base-patch16-224} & ViT-B/16 & 86M & $224^2$ & 196 & 768\\
\textcolor{hiveAmber}{InternViT} & \texttt{OpenGVLab/}\allowbreak\texttt{InternViT-300M-448px-V2\_5} & ViT, patch 14 & 300M & $224^2$ & 256 & 1{,}024\\
\textcolor{hiveAmber}{SigLIP} & \texttt{google/}\allowbreak\texttt{siglip-base-patch16-224} & ViT-B/16 & 86M & $224^2$ & 196 & 768\\
\textcolor{hiveAmber}{CLIP} & \texttt{openai/}\allowbreak\texttt{clip-vit-base-patch16} & ViT-B/16 & 86M & $224^2$ & 196 & 768\\
\textcolor{hiveCyan}{VC-1} & \texttt{facebook/}\allowbreak\texttt{vc1-base} & ViT-B/16 & 86M & $224^2$ & 196 & 768\\
\textcolor{hiveCyan}{Voltron} & \texttt{v-cond-base} (\texttt{voltron-robotics}) & ViT-B/16 & 86M & $224^2$ & 196 & 768\\
\textcolor{hiveIndigo}{Theia} & \texttt{theaiinstitute/}\allowbreak\texttt{}\allowbreak\texttt{theia-base-patch16-224-cdiv} & DeiT-B/16 & 86M & $224^2$ & 196 & 768\\
\textcolor{hiveIndigo}{C-RADIOv4} & \texttt{c-radio\_v4-h} (\texttt{NVlabs/}\allowbreak\texttt{RADIO} hub) & ViT-H/16 & 631M & $224^2$ & 196 & 1{,}280\\
\textcolor{hiveIndigo}{RADIOv2.5} & \texttt{radio\_v2.5-b} (\texttt{NVlabs/}\allowbreak\texttt{RADIO} hub) & ViT-B/16 & 98M & $224^2$ & 196 & 768\\
\textcolor{hiveMuted}{ViT} & \texttt{google/}\allowbreak\texttt{vit-base-patch16-224} & ViT-B/16 & 86M & $224^2$ & 196 & 768\\
\bottomrule[0.8pt]
\end{tabular*}
\endgroup

\vspace{2pt}
{\footnotesize\leftskip0pt\rightskip0pt\parfillskip0pt plus 0.5\linewidth $^{a}$Concatenation of the frame-attention and global-attention outputs of the last aggregator block. $^{b}$Projected to 1{,}024 channels by the VLM-style interface described below; VLM parameters include the language model and the vision tower.}
\end{table}

\paragraph{Preprocessing.}
Default views are stored and loaded at $224\times224$ and then passed through the encoder's own preprocessing. Hugging Face image processors are used for DINOv2, DINOv3, CLIP, SigLIP, SigLIP2, MAE, and ViT. DINOv2's processor resizes the short side to 256 and center-crops 224, so its policy sees the central 87.5\% of each view; CLIP resizes and center-crops to 224, which leaves a $224\times224$ view unchanged; the others resize to $224\times224$. DINOv2, DINOv3, and MAE use ImageNet statistics, CLIP its own, and SigLIP, SigLIP2, and ViT a mean and standard deviation of 0.5. InternViT and VC-1 resize to $224\times224$ with ImageNet statistics, and V-JEPA~2.1 resizes to $384\times384$ with ImageNet statistics and runs in image mode with one temporal slot. VGGT-$\Omega$, SPA, RADIOv2.5, and C-RADIOv4 receive $[0,1]$ images and apply their own internal normalization, Theia receives 8-bit images and normalizes them internally, and Voltron uses its released transform. LingBot-Vision is resized bilinearly to $224\times224$ with ImageNet statistics. The three VLMs use the Qwen3-VL image processor, which enlarges a $224\times224$ view to $256\times256$ to meet its minimum pixel count; 16-pixel patches and $2\times2$ merging then yield 64 image tokens per view.

\paragraph{Token extraction.}
Each encoder receives its prescribed image normalization exactly once. Class and register tokens are removed and only patch tokens are kept. VGGT-$\Omega$ contributes the patch tokens of its last aggregator block and drops the camera and register tokens. On RoboTwin it receives the head view and the two wrist views together as one three-frame sequence, so its global-attention layers exchange information across the views before the policy reads the $3\times196$ tokens; every other encoder processes each view on its own. Its diagnostics, like all others, read only the head camera. MAE runs with a masking ratio of 0; the released implementation still permutes the order of its output tokens, which leaves the policy unaffected because its cross-attention does not depend on token order, but makes MAE's position-dependent statistics meaningless. LingBot-Vision contributes its final normalized patch tokens. For the VLMs, the images of all views are placed in the chat template before the instruction, the model runs once, and the hidden states at the image-token positions of the chosen layer are kept, the final layer unless stated otherwise. The language model attends causally and the image tokens precede the instruction, so these hidden states do not depend on the instruction; image tokens of a later view can attend to those of earlier views. The instruction reaches the policy only through the CLIP text tokens described in Appendix~\ref{app:protocol}, which enter the context after the image tokens.

\begin{table}[!ht]
\centering
\caption{\textbf{Checkpoints of the model-scale study} (Section~\ref{sec:results-model}). Every row is read through the same interface as its default checkpoint in Table~\ref{tab:interfaces}; the default rows are repeated for reference.}
\label{tab:scale-variants}
\begingroup
\arrayrulecolor{hiveInk}
\setlength{\tabcolsep}{3pt}
\renewcommand{\arraystretch}{1.08}
\scriptsize
\begin{tabular*}{\linewidth}{@{\extracolsep{\fill}}llllrrr@{}}
\toprule[0.8pt]
\textbf{Family} & \textbf{Size} & \textbf{Checkpoint} & \textbf{Backbone} & \textbf{Params} & \textbf{$N$} & \textbf{$D$}\\
\midrule[0.45pt]
\multirow{4}{*}{\textcolor{hiveAccent}{DINOv2}} & S & \texttt{facebook/dinov2-small} & ViT-S/14 & 21M & 256 & 384\\
 & B (default) & \texttt{facebook/dinov2-base} & ViT-B/14 & 86M & 256 & 768\\
 & L & \texttt{facebook/dinov2-large} & ViT-L/14 & 300M & 256 & 1{,}024\\
 & g & \texttt{facebook/dinov2-giant} & ViT-g/14 & 1.1B & 256 & 1{,}536\\
\midrule[0.25pt]
\multirow{4}{*}{\textcolor{hiveAccent}{DINOv3}} & S & \texttt{facebook/dinov3-vits16-pretrain-lvd1689m} & ViT-S/16 & 21M & 196 & 384\\
 & B (default) & \texttt{facebook/dinov3-vitb16-pretrain-lvd1689m} & ViT-B/16 & 86M & 196 & 768\\
 & L & \texttt{facebook/dinov3-vitl16-pretrain-lvd1689m} & ViT-L/16 & 300M & 196 & 1{,}024\\
 & H+ & \texttt{facebook/dinov3-vith16plus-pretrain-lvd1689m} & ViT-H+/16 & 840M & 196 & 1{,}280\\
\midrule[0.25pt]
\multirow{3}{*}{\textcolor{hiveAccent}{V-JEPA~2.1}} & B (default) & \texttt{vjepa2\_1\_vit\_base\_384} & ViT-B/16 & 80M & 576 & 768\\
 & L & \texttt{vjepa2\_1\_vit\_large\_384} & ViT-L/16 & 300M & 576 & 1{,}024\\
 & g & \texttt{vjepa2\_1\_vit\_giant\_384} & ViT-g/16 & 1B & 576 & 1{,}408\\
\midrule[0.25pt]
\multirow{2}{*}{\textcolor{hiveCyan}{VC-1}} & B (default) & \texttt{facebook/vc1-base} & ViT-B/16 & 86M & 196 & 768\\
 & L & \texttt{facebook/vc1-large} & ViT-L/16 & 307M & 196 & 1{,}024\\
\midrule[0.25pt]
\bottomrule[0.8pt]
\end{tabular*}
\endgroup
\end{table}

\section{Extended Closed-Loop Results}
\label{sec:results}

This appendix provides the paired results and additional controls behind Section~\ref{sec:m-results}. The complete task-level values appear in Appendix~\ref{app:results}.

\subsection{Frozen encoders and task sharing}
\label{sec:results-frozen}\label{sec:results-sharing}

Table~\ref{tab:suite} compares the 20 default encoders under multi-task training. Table~\ref{tab:sharing} adds the matched single-task runs on RoboCasa. The 17 matched encoders average 26.94\% under ST and 23.62\% under MT. The ratio of these averages is 87.7\%; averaging the individual retention ratios gives 87.8\%. Both summarize the same broad pattern: the reduction in total training steps is much larger than the loss in success.

Each policy receives 100k updates at global batch 128. Covering twelve tasks therefore uses 1.2M optimizer steps for the collection of ST policies and 100k for one MT policy. The twelvefold ratio describes total optimizer steps; MT additionally conditions on language. Table~\ref{tab:defaults} gives the learning rates for each regime.

\begin{table}[!ht]\centering\small
\caption{\textbf{Matched RoboCasa ST and MT results.} Both use 1,000 demonstrations per task. ST comprises twelve independently trained policies, MT one shared policy. $\Delta$ is MT minus ST in percentage points; retained success is $100\,\mathrm{MT}/\mathrm{ST}$.}
\label{tab:sharing}\setlength{\tabcolsep}{7pt}
\begin{tabular}{lrrrr}\toprule
\textbf{Encoder} & \textbf{ST} & \textbf{MT} & $\boldsymbol{\Delta}$ & \textbf{Retained (\%)}\\\midrule
VGGT-$\Omega$ & 43.17{\scriptsize\,$\pm$1.42} & 35.39{\scriptsize\,$\pm$1.33} & -7.78 & 82.0\\
LingBot-Vision & 31.78{\scriptsize\,$\pm$0.92} & 29.39{\scriptsize\,$\pm$1.49} & -2.39 & 92.5\\
DINOv2 & 37.33{\scriptsize\,$\pm$0.94} & 32.28{\scriptsize\,$\pm$1.45} & -5.05 & 86.5\\
DINOv3 & 32.50{\scriptsize\,$\pm$1.41} & 31.22{\scriptsize\,$\pm$1.34} & -1.28 & 96.1\\
MAE & 22.89{\scriptsize\,$\pm$1.13} & 25.28{\scriptsize\,$\pm$1.49} & +2.39 & 110.4\\
V-JEPA~2.1 & 23.39{\scriptsize\,$\pm$1.04} & 18.67{\scriptsize\,$\pm$1.22} & -4.72 & 79.8\\
Xiaomi-Robotics-1 & 24.83{\scriptsize\,$\pm$1.55} & 25.17{\scriptsize\,$\pm$1.70} & +0.34 & 101.4\\
DepthVLM & 28.22{\scriptsize\,$\pm$1.57} & 18.50{\scriptsize\,$\pm$1.08} & -9.72 & 65.6\\
Qwen3-VL & 23.78{\scriptsize\,$\pm$1.69} & 19.56{\scriptsize\,$\pm$1.19} & -4.22 & 82.3\\
SigLIP2 & 29.78{\scriptsize\,$\pm$1.60} & 29.11{\scriptsize\,$\pm$1.53} & -0.67 & 97.8\\
InternViT & 22.39{\scriptsize\,$\pm$1.05} & 20.11{\scriptsize\,$\pm$0.97} & -2.28 & 89.8\\
SigLIP & 16.61{\scriptsize\,$\pm$1.12} & 13.67{\scriptsize\,$\pm$1.28} & -2.94 & 82.3\\
CLIP & 15.17{\scriptsize\,$\pm$1.52} & 11.44{\scriptsize\,$\pm$0.98} & -3.73 & 75.4\\
VC-1 & 32.61{\scriptsize\,$\pm$1.16} & 28.44{\scriptsize\,$\pm$1.42} & -4.17 & 87.2\\
Voltron & 31.94{\scriptsize\,$\pm$1.25} & 24.83{\scriptsize\,$\pm$1.01} & -7.11 & 77.7\\
Theia & 27.00{\scriptsize\,$\pm$1.29} & 24.83{\scriptsize\,$\pm$1.08} & -2.17 & 92.0\\
ViT & 14.67{\scriptsize\,$\pm$1.09} & 13.67{\scriptsize\,$\pm$1.23} & -1.00 & 93.2\\
\bottomrule\end{tabular}\end{table}

\subsection{Patch tokens and fine-tuning}
\label{sec:results-gap}\label{sec:results-ft}

\begin{figure}[!ht]\centering
\includegraphics[width=\linewidth]{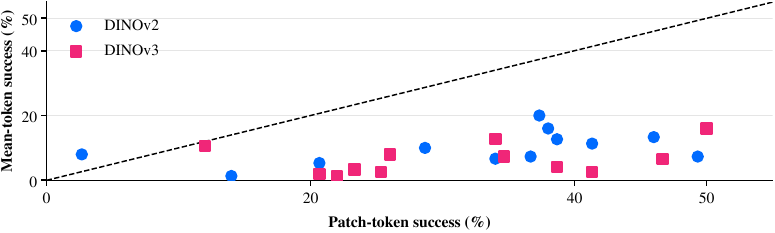}
\caption{\textbf{Spatial pooling on individual RoboCasa tasks.} Each point is one encoder--task pair. Points below the diagonal lose success when patch tokens are replaced by their mean.}
\label{fig:pooling-tasks}\end{figure}

The mean-pooling control uses frozen DINOv2 and DINOv3 on RoboCasa MT, replacing 256 and 196 tokens per view, respectively, with one mean token. All other policy and optimization settings are held fixed. Figure~\ref{fig:m-closedloop} reports suite means; Figure~\ref{fig:pooling-tasks} shows the individual tasks. Pooling lowers success in 23 of the 24 pairs. This control evaluates mean pooling rather than learned compression.

Table~\ref{tab:finetune} gives the five frozen--fine-tuned pairs on each benchmark. The Spearman correlation between frozen and adapted success is 0.70 on RoboCasa and 1.00 on RoboTwin. Fine-tuning uses an encoder learning rate equal to one tenth of the action-head rate.

\begin{table}[!ht]\centering\footnotesize
\caption{\textbf{Frozen and fine-tuned success.} $\Delta$ is the gain in percentage points.}
\label{tab:finetune}
\begin{tabular*}{\linewidth}{@{\extracolsep{\fill}}lrrr rrr@{}}\toprule
&\multicolumn{3}{c}{\textbf{RoboCasa ST}}&\multicolumn{3}{c}{\textbf{RoboTwin MT}}\\
\cmidrule(lr){2-4}\cmidrule(lr){5-7}
\textbf{Encoder}&\textbf{Frozen}&\textbf{FT}&$\boldsymbol{\Delta}$&\textbf{Frozen}&\textbf{FT}&$\boldsymbol{\Delta}$\\\midrule
VGGT-$\Omega$ & 43.17{\scriptsize\,$\pm$1.42} & 50.94{\scriptsize\,$\pm$1.33} & +7.77 & 50.00{\scriptsize\,$\pm$1.33} & 76.06{\scriptsize\,$\pm$1.16} & +26.06\\
DINOv2 & 37.33{\scriptsize\,$\pm$0.94} & 39.72{\scriptsize\,$\pm$1.74} & +2.39 & 49.39{\scriptsize\,$\pm$1.04} & 51.61{\scriptsize\,$\pm$1.29} & +2.22\\
DINOv3 & 32.50{\scriptsize\,$\pm$1.41} & 41.94{\scriptsize\,$\pm$1.46} & +9.44 & 30.44{\scriptsize\,$\pm$1.44} & 48.61{\scriptsize\,$\pm$1.33} & +18.17\\
VC-1 & 32.61{\scriptsize\,$\pm$1.16} & 38.78{\scriptsize\,$\pm$1.29} & +6.17 & 25.06{\scriptsize\,$\pm$1.46} & 44.00{\scriptsize\,$\pm$1.22} & +18.94\\
SigLIP2 & 29.78{\scriptsize\,$\pm$1.60} & 30.89{\scriptsize\,$\pm$1.30} & +1.11 & 44.22{\scriptsize\,$\pm$1.24} & 48.72{\scriptsize\,$\pm$1.41} & +4.50\\
\bottomrule\end{tabular*}\end{table}

\subsection{Demonstration and model scale}
\label{sec:results-data}\label{sec:results-model}

\begin{figure}[!ht]\centering
\includegraphics[width=\linewidth]{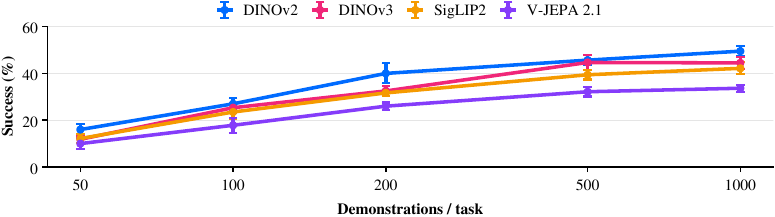}
\caption{\textbf{RoboCasa ST demonstration scaling on C1--C4.} Each task has its own policy. The 1,000-demonstration endpoint averages the same four tasks from the main ST runs. Error bars propagate the four task deviations.}
\label{fig:data-scale}\end{figure}

The multi-task sweeps in Figure~\ref{fig:m-scale} average all twelve tasks. The RoboCasa ST sweep averages only C1--C4 (Figure~\ref{fig:data-scale}), so comparisons between ST and MT use those four tasks on both sides. Smaller demonstration sets are nested subsets of the larger sets. The full-data endpoints reuse the corresponding main experiments.

The model-size sweep covers DINOv2, DINOv3, V-JEPA~2.1, and VC-1. The nine additional checkpoints and their token interfaces are listed in Table~\ref{tab:scale-variants}; their task-level results appear in Appendix~\ref{app:results}. Default checkpoints are marked by rings in Figure~\ref{fig:m-scale}.

\subsection{State and image-resolution controls}
\label{sec:results-inputs}

The state controls add one token from a two-layer $\mathrm{input}\to768\to768$ ReLU MLP. RoboCasa encodes 29 joint coordinates as 58 sine/cosine features; RoboTwin normalizes 12 joints and retains two gripper indicators. DINOv2 and RADIO improve on both suites without reversing order. This pair tests whether absent proprioception explains their gap; it cannot establish how state/IDM correlations change across the full encoder set. Section~\ref{sec:m-diag} correlates diagnostics with the default image-only policies. The resolution control resamples stored $224\times224$ images to $320\times320$, increasing DINOv3's patches from 196 to 400 per view. It adds tokens, not newly captured image detail. Despite exceeding DINOv2's 256 tokens, RoboTwin success remains 14.78 points lower.

\begin{table}[!ht]\centering\footnotesize
\caption{\textbf{Input and action-head ablations.} Baseline is the corresponding frozen MT policy. Entries preserve the exported suite evaluation deviations; these new ablations do not all use the task-error propagation convention of the main comparison. $\Delta$ denotes the change in suite mean.}
\label{tab:input-ablations}\setlength{\tabcolsep}{3pt}
\begin{tabular*}{\linewidth}{@{\extracolsep{\fill}}lllrrr@{}}\toprule
\textbf{Suite}&\textbf{Encoder}&\textbf{Change}&\textbf{Baseline}&\textbf{Ablated}&$\boldsymbol{\Delta}$\\\midrule
RoboCasa & DINOv2 & + state & 32.28{\scriptsize\,$\pm$1.45} & 35.25{\scriptsize\,$\pm$2.98} & +2.97\\
RoboCasa & RADIOv2.5 & + state & 24.39{\scriptsize\,$\pm$1.11} & 26.17{\scriptsize\,$\pm$1.20} & +1.78\\
RoboCasa & DINOv3 & $320\times320$ & 31.22{\scriptsize\,$\pm$1.34} & 32.39{\scriptsize\,$\pm$1.34} & +1.17\\
RoboCasa & DINOv2 & No query tokens & 32.28{\scriptsize\,$\pm$1.45} & 27.78{\scriptsize\,$\pm$2.68} & -4.50\\
RoboCasa & DINOv3 & No query tokens & 31.22{\scriptsize\,$\pm$1.34} & 23.50{\scriptsize\,$\pm$1.69} & -7.72\\
RoboTwin & DINOv2 & + state & 49.39{\scriptsize\,$\pm$1.04} & 49.78{\scriptsize\,$\pm$2.04} & +0.39\\
RoboTwin & RADIOv2.5 & + state & 31.11{\scriptsize\,$\pm$1.19} & 33.06{\scriptsize\,$\pm$2.64} & +1.95\\
RoboTwin & DINOv3 & $320\times320$ & 30.44{\scriptsize\,$\pm$1.44} & 34.61{\scriptsize\,$\pm$2.00} & +4.17\\
\bottomrule\end{tabular*}\end{table}

\subsection{Learned query tokens in the action head}
\label{sec:results-register}

The default DiT prepends 32 learned query tokens to the 16 noisy action tokens. These queries supply additional latent positions through which self-attention and cross-attention can exchange information; the action decoder reads the action-token positions. They are part of the action head, distinct from class or register tokens produced by an encoder.

Removing the 32 queries reduces RoboCasa MT success from 32.28\% to 27.78\% for DINOv2 and from 31.22\% to 23.50\% for DINOv3 (Table~\ref{tab:input-ablations}). Both policies still learn, and DINOv2 remains ahead, but the extra latent positions improve success by 4.50 and 7.72 points. The learned queries help the common head use visual context.

\subsection{The extraction layer}
\label{sec:results-layer}
\begin{figure}[!ht]\centering
\includegraphics[width=\linewidth]{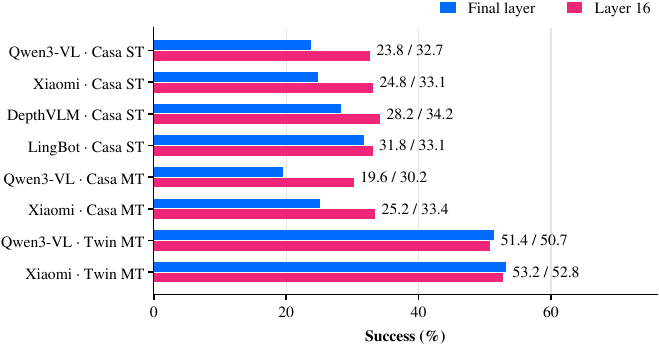}
\caption{\textbf{Final-layer features versus layer 16.} Labels give final-layer / layer-16 success. Six settings use RoboCasa, two RoboTwin; LingBot-Vision provides a vision-encoder layer control.}
\label{fig:vlm-layers}\end{figure}

On RoboCasa ST, layer 16 improves Qwen3-VL by 8.89 points, Xiaomi-Robotics-1 by 8.28, and DepthVLM by 5.95. The MT gains are 10.66 points for Qwen3-VL and 8.27 for Xiaomi-Robotics-1. LingBot-Vision gains 1.28 points on RoboCasa ST. On RoboTwin, Qwen3-VL and Xiaomi-Robotics-1 change by $-0.72$ and $-0.44$ points. The value of an intermediate layer therefore depends on the downstream setting; the layer is part of the representation choice.

The paired readouts also distinguish information recovery from policy success. On RoboTwin, layer 16 lowers Qwen3-VL's IDM error from 0.177 to 0.124 and Xiaomi's from 0.150 to 0.125; both state errors fall as well, without a corresponding success gain. On RoboCasa, Xiaomi's IDM error instead rises slightly (0.158 to 0.168) while MT success gains 8.27 points. A probe's ordering across default encoders therefore need not predict the effect of changing the extraction layer within one encoder.

\subsection{Variation across training seeds}
\label{sec:results-seeds}

We repeat RoboCasa MT training for four encoders with seeds 1, 42, and 78 (Table~\ref{tab:train-seeds}). All three runs give the same ordering: DINOv2, VC-1, RADIOv2.5, then V-JEPA~2.1. Across-training standard deviations are 1.74 points for DINOv2, 0.18 for RADIOv2.5, 1.10 for VC-1, and 0.93 for V-JEPA~2.1. The tested multi-task policies are thus broadly stable across training seeds: aggregate success changes modestly, and the encoder ordering is preserved.

DINOv2 varies most, from 28.22\% to 32.28\%, so its single-seed score is best read alongside the three-run mean of 29.87\%. The table separates training variability from the rollout variability reported elsewhere.

\begin{table}[!ht]\centering\small
\caption{\textbf{Training-seed variation on RoboCasa MT.} Each seed column contains its twelve-task mean success, averaged over three evaluation seeds. The last two columns summarize the three training runs; SD is the population standard deviation across training seeds.}
\label{tab:train-seeds}
\begin{tabular}{lrrrrr}\toprule
\textbf{Encoder}&\textbf{Seed 1}&\textbf{Seed 42}&\textbf{Seed 78}&\textbf{Mean}&\textbf{Training SD}\\\midrule
DINOv2 & 28.22 & 32.28 & 29.11 & 29.87 & 1.74\\
RADIOv2.5 & 24.56 & 24.39 & 24.83 & 24.59 & 0.18\\
VC-1 & 25.78 & 28.44 & 26.78 & 27.00 & 1.10\\
V-JEPA~2.1 & 19.94 & 18.67 & 20.94 & 19.85 & 0.93\\
\bottomrule\end{tabular}\end{table}

\section{Physical-Robot Evaluation}
\label{app:realworld}

\paragraph{DINOv2's lead carries over to three physical manipulation tasks.}
We compare DINOv2, DINOv3, and CLIP on a real bimanual platform to test whether their simulation ordering survives changes in appearance, sensing, and execution. DINOv2 leads on every task, averaging 49.67\% success versus 41.00\% for DINOv3 and 17.00\% for CLIP (Table~\ref{tab:realworld}). This ordering agrees with both simulation suites. The comparison extends that result to physical interaction; the wider diagnostic relationships remain questions for a larger real-robot encoder study.

\begin{figure}[!ht]\centering
\includegraphics[width=\linewidth]{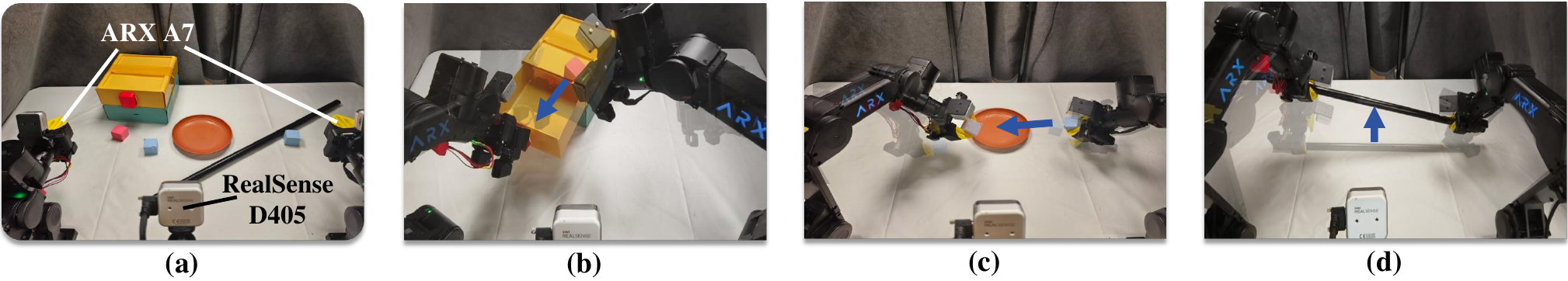}
\caption{\textbf{Physical platform and tasks.} (a) The ARX A7 bimanual robot and egocentric RealSense D405 camera. (b) Drawer Cube Place. (c) Pick Two Cube. (d) Grab Stick.}
\label{fig:realworld}\end{figure}

\paragraph{Platform and tasks.}
The ARX A7 has 14 action coordinates across its two arms and grippers. An Intel RealSense D405 supplies the egocentric RGB view, resized to $224\times224$. Drawer Cube Place requires opening a drawer and placing a cube inside. Pick Two Cube requires grasping two cubes, one with each gripper, and placing both on a plate. Grab Stick requires aligning the grippers with a slender rod. These tasks test sequential interaction, bimanual coordination, and precise alignment (Figure~\ref{fig:realworld}).

\paragraph{Training and evaluation.}
We use 50 demonstrations per task and a shared multi-task training set. Policies are trained for 20,000 updates at batch size 32 and predict 16-step chunks of absolute joint positions. The real-robot configuration includes RGB, a task instruction, and 14-dimensional robot state, with four Euler steps at inference. Each encoder is evaluated in 100 trials per task, giving 300 trials per encoder. Table~\ref{tab:realworld} reports success percentages and their equally weighted mean.

\begin{table}[!ht]\centering\small
\caption{\textbf{Real-robot success (\%).} Each entry uses 100 trials; the mean weights the three tasks equally. DINOv2 leads throughout, with its largest DINOv2--DINOv3 gap on Grab Stick.}
\label{tab:realworld}\setlength{\tabcolsep}{6pt}
\begin{tabular*}{\linewidth}{@{\extracolsep{\fill}}lrrrr@{}}\toprule
\rowcolor{hiveAccent!9}Encoder & Drawer Cube Place & Pick Two Cube & Grab Stick & Mean\\\midrule
DINOv2 & \best{42} & \best{31} & \best{76} & \best{49.67}\\
DINOv3 & 32 & 27 & 64 & 41.00\\
CLIP & 17 & 11 & 23 & 17.00\\\bottomrule
\end{tabular*}\end{table}

\paragraph{The advantage spans different stages of manipulation.}
DINOv2 exceeds DINOv3 by 10 points on Drawer Cube Place, 4 on Pick Two Cube, and 12 on Grab Stick. All three encoders perform best on Grab Stick, so the ordering is shared even as task difficulty changes. CLIP falls behind both DINO variants on every task. These results favor the same encoder across three different physical demands, rather than through one unusually favorable task.
\clearpage

\section{Diagnostics: Definitions and Complete Analyses}
\label{app:diagnostics}\label{app:metrics}\label{sec:diag-defs}

\subsection{Token measurements}
For frame $t$, let $z_{t,i}\in\mathbb R^D$ be patch $i$ on an $H\times W$ grid, $r_{t,i}=\|z_{t,i}\|_2$, and $g_t=N^{-1}\sum_i z_{t,i}$. We sample every fifth demonstration, frames 0--100 at stride 5, using the head camera. Frame-level statistics are averaged within a video, then across videos and tasks. This describes the early portion of each demonstration. VLM tokens are included in readouts but not the grid-based token comparison. MAE's output order is permuted, so neighbor, frequency, patch-temporal, and temporal norm-histogram statistics omit MAE.

\paragraph{Spatial statistics.}
Mean token cosine averages $\cos(z_i,z_j)$ over $i\ne j$; neighbor similarity restricts this average to horizontal and vertical grid neighbors. Mean norm averages $r_i$, and token-to-global cosine averages $\cos(z_i,g)$. Effective rank is the participation ratio of the centered patch covariance $C$, with eigenvalues $\lambda_k$:
\begin{equation}
\mathrm{PR}(C)=(\textstyle\sum_k\lambda_k)^2/\textstyle\sum_k\lambda_k^2.
\end{equation}
Norm-histogram entropy bins $r_i$ into 30 equal-width bins over that frame's own minimum and maximum, then computes $-\sum_b q_b\ln q_b$. It lies in $[0,\ln30]$. Identical norms occupy one bin; a few exceptionally large norms can also lower entropy by compressing the remaining norms into a few bins. It is a histogram-shape statistic, rather than a measure of uniform attention over patches.

\paragraph{Temporal statistics.}
Temporal smoothness is the mean drift $(T-1)^{-1}\sum_t\|g_{t+1}-g_t\|_2$; smaller values mean less change. Temporal cosine shift averages $1-\cos(g_t,g_{t+1})$. Temporal variance averages $\mathrm{Var}_t(g_{t,d})$ over channels. Effective rank and spectral entropy use the covariance eigenvalues of the time-centered sequence of pooled features; entropy is $-\sum_jp_j\ln p_j$, with $p_j=\lambda_j/\sum_k\lambda_k$. This temporal ``spectral'' entropy refers to covariance eigenvalues, not a Fourier spectrum. Lag-one autocorrelation averages $\sum_{t=1}^{T-1}x_{t,d}x_{t+1,d}/\sum_{t=1}^T x_{t,d}^2$ over nonconstant centered channels $x_{t,d}=g_{t,d}-\bar g_d$.

Trajectory total variation divides the summed drift by $T$, making it a near-duplicate of temporal smoothness. Patch temporal smoothness averages $\|z_{t+1,i}-z_{t,i}\|_2$ over time and corresponding patch positions. Temporal norm entropy uses 16 equal-width bins over the range of each position's norms through a video, divides the histogram entropy by $\ln16$, and averages over positions. The trajectory variance ratio divides mean within-video temporal variance by the variance of video means, averaged over channels, with a small numerical stabilizer in the denominator.

\paragraph{Spatial frequency.}
Each token channel is reshaped to its patch grid, centered spatially, and transformed with an orthonormal 2D Fourier transform. Normalized power $p_d(u,v)$ sums to one per channel. With frequency radius $\varrho$ normalized by the grid's maximum radius, low, mid, and high shares sum power over $\varrho\le0.15$, $0.15<\varrho\le0.35$, and $\varrho>0.35$. Spectral entropy is $-\sum p_d\ln p_d/\ln(HW)$, centroid is $\mu_d=\sum p_d\varrho$, and bandwidth is $\sqrt{\sum p_d(\varrho-\mu_d)^2}$. Each is averaged over channels. These summaries describe a shared spectrum and are statistically dependent. Absolute norms, drift, and variance depend on feature scale; cosine, participation ratio, adaptive histogram entropy, and normalized spectra remove uniform scaling.

\subsection{Readouts and controls}
\label{app:probes}
IDM and state readouts use a $D\to256$ input projection, a four-layer pre-norm transformer with four attention heads and feed-forward width 512, and a learned query. The readout applies LayerNorm and a two-layer width-256 head with GELU. Dropout is 0.1. Parameters in the transformer are shared in architecture, not in learned weights, across encoders; only the input projection grows with encoder width. This removes the quadratic growth in state-probe capacity that would occur if the transformer used native feature width.

The two-frame input is $[q;Z_t;Z_{t+16}]$, with within-frame sinusoidal positions and frame-segment embeddings. The single-frame control is $[q;Z_t]$. Both predict the same standardized target $a_{t+15}-s_t$ and report mean squared error per coordinate. The state probe predicts standardized $s_t$. Targets use training-set statistics. Input length changes from $1+N$ to $1+2N$ in the two-frame comparison, so its error reduction combines a later observation with additional attention computation; it does not isolate motion sensitivity.

Probes train for four epochs using AdamW, learning rate $5\times10^{-4}$, weight decay $10^{-4}$, cosine decay to 1\% of the peak, batch 256, clipping at 1, and seed 42. The episode split is 1,000/500 train/held-out on RoboCasa and 400/100 on RoboTwin, stratified across tasks. The lowest held-out error selects the epoch and is the reported value, making it a validation error. Diagnostic encoders remain frozen even when the checkpoint was previously fine-tuned with the policy.

The object probe reads the first head-camera frame and predicts the camera-frame 3D location of the task object on four RoboTwin tasks: bottle, hammer/block, alarm clock, and laptop. It uses a native-width four-layer transformer and reports coordinate-averaged absolute error in meters, with 400 training and 100 held-out episodes per task. Its capacity and task coverage differ from the width-256 action/state probes. All readouts use the head view; RoboTwin policies additionally see wrist views and, for some interfaces, features that mix views.

\subsection{Statistics and analysis cohorts}
\label{app:stats}\label{app:manifest}
Spearman correlation is Pearson correlation between mid-ranks, preserving ties. A two-sided permutation probability uses 100,000 permutations of success across encoders, with $(1+\#\{|\rho_b|\ge|\rho|\})/(100001)$. Cohorts of at most eight use all permutations. Bootstrap intervals use the 2.5th and 97.5th percentiles of 10,000 encoder resamples; resamples with zero rank variance are discarded. Permutations use a fixed seed indexed by sample size, and bootstrap seeds use a CRC-32 hash of the analysis name.

The primary family covers all corrected token/readout metrics: 24 on RoboCasa and 25 on RoboTwin (including object error). Benjamini--Hochberg correction is applied within each cohort. Single-frame IDM is analyzed as a control, with a separate three-readout correction available alongside the primary results. The vision-only cohorts contain 14 encoders on Casa ST and 17 on each MT suite; adding VLM readouts gives 17 and 20. Position-dependent statistics have one fewer encoder because MAE is excluded. The same RoboCasa diagnostic measurements are paired with ST and MT success. Both settings therefore share encoders and diagnostic inputs.

Family deletion removes all encoders in one of the seven groups of Table~\ref{tab:encoder_families} before recomputing $\rho$. The 12 task-level associations reuse the same suite-level diagnostic and correlate it with each task's success. Encoders can also share teachers across these groups, so these checks assess sensitivity to composition rather than establish independent replications. Partial Spearman correlation regresses the ranks of IDM error and success separately on state-error ranks, then correlates the residuals.

Paired adaptation analyses use DINOv2, DINOv3, VC-1, SigLIP2, and VGGT-$\Omega$, each with corrected frozen and fine-tuned success scores. Pooled correlations combine these with the frozen vision cohort; each adapted checkpoint shares its initialization with a frozen one. VGGT's unchanged frequency entries are excluded from frequency-change analyses. No policy-fine-tuned vision-benchmark scores are used.

\paragraph{Family-held-out prediction.}
For each outer fold, ridge regression holds out all encoders in one pretraining family. Each feature is standardized using only the training fold. Inner family-held-out validation selects the penalty from $\{0.01,0.1,1,10,100\}$ by mean absolute error; the intercept is unpenalized. Predictions are evaluated in success points. All models use the same complete-case cohort, excluding VLM token statistics and MAE's temporal norm entropy. This gives 13 encoders for Casa ST and 16 for each MT suite. The mean baseline predicts the outer training mean.

\paragraph{Candidate selection.}
A $k$-candidate shortlist takes the lowest probe errors, then reports the highest observed success among those candidates. Regret is the difference from the best in the full eligible cohort. For success values sorted as $y_{(1)}\le\cdots\le y_{(n)}$, uniform sampling of $k$ distinct candidates has expected best success
\begin{equation}
\mathbb E[\max y]=\sum_{j=k}^n y_{(j)}\binom{j-1}{k-1}/\binom{n}{k}.
\end{equation}
The comparison counts policy-training runs, not total wall time, and uses the full frozen suite including VLMs. It is a retrospective screen on these benchmarks.

\subsection{All metric associations}
\label{tab:diag-assoc}
Tables~\ref{tab:all-CST}--\ref{tab:all-TMT} give all primary associations and the single-frame control. The RoboCasa frequency summaries are more consistent than its readout ordering, whereas IDM is the clearest corrected association on RoboTwin. Mean token cosine and effective rank show little relationship with success across all three settings. The main text separately reports the VLM readout comparison because those tokens do not share the vision encoders' spatial interface.
\begin{table}[!htp]\centering\footnotesize
\caption{\textbf{RoboCasa ST: all diagnostic associations.} Vision encoders; CI is the 95\% bootstrap interval. Sample sizes, permutation tests, and corrections follow Section~\ref{app:stats}.}
\label{tab:all-CST}
\setlength{\tabcolsep}{3pt}\renewcommand{\arraystretch}{1.05}
\begin{tabular*}{\linewidth}{@{\extracolsep{\fill}}lrrrrr@{}}\toprule
Measurement & $n$ & $\rho$ & 95\% CI & $p$ & $q$\\\midrule
Mean token cosine & 14 & +0.14 & [-0.50,+0.70] & 0.638 & 0.696\\
Mean token norm & 14 & -0.08 & [-0.72,+0.53] & 0.785 & 0.819\\
Neighbor similarity & 13 & +0.53 & [-0.04,+0.87] & 0.066 & 0.198\\
Effective rank & 14 & -0.05 & [-0.64,+0.57] & 0.882 & 0.882\\
Norm-histogram entropy & 14 & +0.66 & [+0.13,+0.93] & 0.013 & 0.061\\
Token-to-global cosine & 14 & +0.16 & [-0.48,+0.70] & 0.582 & 0.665\\
Two-frame IDM & 14 & -0.38 & [-0.83,+0.32] & 0.175 & 0.346\\
Robot state & 14 & -0.44 & [-0.76,+0.24] & 0.116 & 0.252\\
Temporal smoothness & 14 & -0.17 & [-0.84,+0.51] & 0.552 & 0.662\\
Temporal cosine shift & 14 & -0.18 & [-0.73,+0.42] & 0.534 & 0.662\\
Temporal variance & 14 & -0.24 & [-0.77,+0.42] & 0.418 & 0.603\\
Temporal effective rank & 14 & -0.58 & [-0.92,+0.01] & 0.032 & 0.114\\
Temporal spectral entropy & 14 & -0.58 & [-0.92,+0.02] & 0.033 & 0.114\\
Autocorrelation & 14 & +0.23 & [-0.41,+0.73] & 0.427 & 0.603\\
Trajectory total variation & 14 & -0.17 & [-0.83,+0.50] & 0.552 & 0.662\\
Patch temporal smoothness & 13 & -0.39 & [-0.82,+0.15] & 0.187 & 0.346\\
Temporal norm entropy & 13 & +0.26 & [-0.36,+0.78] & 0.395 & 0.603\\
Trajectory variance ratio & 14 & -0.35 & [-0.76,+0.28] & 0.227 & 0.389\\
Low-frequency fraction & 13 & \textbf{+0.74} & [+0.36,+0.88] & 0.005 & 0.029\\
Mid-frequency fraction & 13 & +0.48 & [-0.08,+0.84] & 0.096 & 0.231\\
High-frequency fraction & 13 & \textbf{-0.75} & [-0.89,-0.39] & 0.004 & 0.029\\
Spatial spectral entropy & 13 & \textbf{-0.75} & [-0.91,-0.35] & 0.004 & 0.029\\
Spectral centroid & 13 & \textbf{-0.75} & [-0.91,-0.37] & 0.004 & 0.029\\
Spectral bandwidth & 13 & -0.50 & [-0.89,+0.12] & 0.085 & 0.226\\
Single-frame IDM & 14 & -0.42 & [-0.74,+0.20] & 0.133 & --\\
\bottomrule\end{tabular*}\end{table}

\begin{table}[!htp]\centering\footnotesize
\caption{\textbf{RoboCasa MT: all diagnostic associations.} Vision encoders; CI is the 95\% bootstrap interval. Sample sizes, permutation tests, and corrections follow Section~\ref{app:stats}.}
\label{tab:all-CMT}
\setlength{\tabcolsep}{3pt}\renewcommand{\arraystretch}{1.05}
\begin{tabular*}{\linewidth}{@{\extracolsep{\fill}}lrrrrr@{}}\toprule
Measurement & $n$ & $\rho$ & 95\% CI & $p$ & $q$\\\midrule
Mean token cosine & 17 & -0.01 & [-0.55,+0.53] & 0.956 & 0.979\\
Mean token norm & 17 & +0.10 & [-0.52,+0.61] & 0.700 & 0.934\\
Neighbor similarity & 16 & +0.48 & [-0.06,+0.83] & 0.061 & 0.209\\
Effective rank & 17 & +0.13 & [-0.40,+0.60] & 0.622 & 0.909\\
Norm-histogram entropy & 17 & +0.42 & [-0.08,+0.78] & 0.097 & 0.261\\
Token-to-global cosine & 17 & +0.07 & [-0.46,+0.56] & 0.800 & 0.962\\
Two-frame IDM & 17 & -0.47 & [-0.85,+0.09] & 0.056 & 0.209\\
Robot state & 17 & -0.39 & [-0.76,+0.15] & 0.117 & 0.261\\
Temporal smoothness & 17 & +0.05 & [-0.51,+0.60] & 0.842 & 0.962\\
Temporal cosine shift & 17 & +0.01 & [-0.50,+0.50] & 0.979 & 0.979\\
Temporal variance & 17 & +0.02 & [-0.51,+0.59] & 0.946 & 0.979\\
Temporal effective rank & 17 & -0.39 & [-0.77,+0.17] & 0.120 & 0.261\\
Temporal spectral entropy & 17 & -0.41 & [-0.78,+0.15] & 0.102 & 0.261\\
Autocorrelation & 17 & +0.20 & [-0.37,+0.68] & 0.435 & 0.745\\
Trajectory total variation & 17 & +0.05 & [-0.50,+0.60] & 0.842 & 0.962\\
Patch temporal smoothness & 16 & -0.17 & [-0.67,+0.37] & 0.523 & 0.836\\
Temporal norm entropy & 16 & +0.33 & [-0.23,+0.77] & 0.206 & 0.412\\
Trajectory variance ratio & 17 & -0.12 & [-0.55,+0.41] & 0.644 & 0.909\\
Low-frequency fraction & 16 & \textbf{+0.68} & [+0.24,+0.89] & 0.005 & 0.030\\
Mid-frequency fraction & 16 & +0.25 & [-0.29,+0.66] & 0.358 & 0.660\\
High-frequency fraction & 16 & \textbf{-0.70} & [-0.89,-0.28] & 0.003 & 0.030\\
Spatial spectral entropy & 16 & \textbf{-0.68} & [-0.92,-0.22] & 0.005 & 0.030\\
Spectral centroid & 16 & \textbf{-0.68} & [-0.91,-0.22] & 0.004 & 0.030\\
Spectral bandwidth & 16 & -0.51 & [-0.89,+0.09] & 0.045 & 0.209\\
Single-frame IDM & 17 & -0.50 & [-0.79,+0.00] & 0.040 & --\\
\bottomrule\end{tabular*}\end{table}

\begin{table}[!htp]\centering\footnotesize
\caption{\textbf{RoboTwin MT: all diagnostic associations.} Vision encoders; CI is the 95\% bootstrap interval. Sample sizes, permutation tests, and corrections follow Section~\ref{app:stats}.}
\label{tab:all-TMT}
\setlength{\tabcolsep}{3pt}\renewcommand{\arraystretch}{1.05}
\begin{tabular*}{\linewidth}{@{\extracolsep{\fill}}lrrrrr@{}}\toprule
Measurement & $n$ & $\rho$ & 95\% CI & $p$ & $q$\\\midrule
Mean token cosine & 17 & -0.15 & [-0.62,+0.41] & 0.560 & 0.752\\
Mean token norm & 17 & +0.42 & [-0.14,+0.80] & 0.099 & 0.562\\
Neighbor similarity & 16 & +0.20 & [-0.30,+0.64] & 0.447 & 0.672\\
Effective rank & 17 & +0.06 & [-0.42,+0.49] & 0.824 & 0.930\\
Norm-histogram entropy & 17 & +0.26 & [-0.25,+0.66] & 0.311 & 0.625\\
Token-to-global cosine & 17 & -0.14 & [-0.60,+0.42] & 0.593 & 0.752\\
Two-frame IDM & 17 & \textbf{-0.75} & [-0.90,-0.40] & $<0.001$ & 0.020\\
Robot state & 17 & -0.62 & [-0.87,-0.17] & 0.009 & 0.112\\
Object position & 17 & -0.54 & [-0.87,+0.01] & 0.027 & 0.226\\
Temporal smoothness & 17 & +0.33 & [-0.24,+0.81] & 0.202 & 0.562\\
Temporal cosine shift & 17 & +0.24 & [-0.32,+0.69] & 0.355 & 0.635\\
Temporal variance & 17 & +0.39 & [-0.24,+0.85] & 0.118 & 0.562\\
Temporal effective rank & 17 & +0.02 & [-0.59,+0.56] & 0.937 & 0.937\\
Temporal spectral entropy & 17 & +0.03 & [-0.55,+0.57] & 0.900 & 0.937\\
Autocorrelation & 17 & -0.05 & [-0.54,+0.43] & 0.855 & 0.930\\
Trajectory total variation & 17 & +0.33 & [-0.25,+0.82] & 0.199 & 0.562\\
Patch temporal smoothness & 16 & +0.20 & [-0.35,+0.67] & 0.457 & 0.672\\
Temporal norm entropy & 16 & +0.26 & [-0.22,+0.66] & 0.325 & 0.625\\
Trajectory variance ratio & 17 & +0.21 & [-0.41,+0.80] & 0.409 & 0.672\\
Low-frequency fraction & 16 & +0.34 & [-0.19,+0.71] & 0.202 & 0.562\\
Mid-frequency fraction & 16 & -0.13 & [-0.64,+0.39] & 0.626 & 0.752\\
High-frequency fraction & 16 & -0.29 & [-0.68,+0.24] & 0.265 & 0.603\\
Spatial spectral entropy & 16 & -0.34 & [-0.72,+0.19] & 0.198 & 0.562\\
Spectral centroid & 16 & -0.31 & [-0.70,+0.25] & 0.232 & 0.579\\
Spectral bandwidth & 16 & -0.13 & [-0.64,+0.48] & 0.632 & 0.752\\
Single-frame IDM & 17 & -0.65 & [-0.90,-0.19] & 0.006 & --\\
\bottomrule\end{tabular*}\end{table}

\clearpage

\subsection{Adaptation and candidate selection}
\begin{table}[!htp]\centering\footnotesize
\caption{\textbf{Diagnostic associations through RoboTwin fine-tuning.} Pre: frozen metric versus FT success; Post: FT metric versus FT success; $\Delta$: metric change versus success change; Pooled: frozen vision encoders plus paired FT checkpoints. ``Falls'' counts lower metric values after FT. Signs are raw, so negative errors are favorable; these small-sample comparisons are exploratory.}
\label{tab:adaptation}\setlength{\tabcolsep}{3pt}
\begin{tabular*}{\linewidth}{@{\extracolsep{\fill}}lrrrrrr@{}}\toprule
Metric & $n$ & Pre & Post & $\Delta$ & Pooled & Falls\\\midrule
Mean token cosine & 5 & +0.10 & -0.60 & -0.90 & -0.42 & 5/5\\
Mean token norm & 5 & +0.80 & +0.60 & -0.10 & +0.23 & 5/5\\
Neighbor similarity & 5 & +0.20 & -0.80 & -0.70 & +0.10 & 2/5\\
Effective rank & 5 & -0.20 & -0.30 & +0.80 & -0.08 & 3/5\\
Norm-histogram entropy & 5 & -0.20 & -0.30 & -1.00 & +0.27 & 2/5\\
Token-to-global cosine & 5 & +0.10 & -0.60 & -0.70 & -0.39 & 5/5\\
Two-frame IDM & 5 & -0.10 & -0.40 & -0.60 & -0.80 & 5/5\\
Robot state & 5 & +0.50 & -0.60 & -0.40 & -0.70 & 5/5\\
Object position & 5 & -0.90 & +0.60 & +0.40 & -0.43 & 1/5\\
Temporal smoothness & 5 & +0.90 & +0.80 & +0.40 & +0.21 & 4/5\\
Temporal cosine shift & 5 & +0.40 & -0.60 & +0.90 & +0.40 & 1/5\\
Temporal variance & 5 & +1.00 & +0.60 & -0.30 & +0.43 & 3/5\\
Temporal effective rank & 5 & +0.30 & -0.40 & +0.90 & -0.15 & 3/5\\
Temporal spectral entropy & 5 & +0.30 & -0.30 & +0.80 & -0.20 & 3/5\\
Autocorrelation & 5 & -0.60 & -0.20 & -0.60 & +0.37 & 0/5\\
Trajectory total variation & 5 & +0.90 & +0.80 & +0.40 & +0.21 & 4/5\\
Patch temporal smoothness & 5 & +0.50 & +0.80 & +0.80 & +0.03 & 4/5\\
Temporal norm entropy & 5 & -0.40 & +0.10 & +0.70 & -0.03 & 4/5\\
Trajectory variance ratio & 5 & +0.90 & -0.20 & +0.20 & +0.54 & 0/5\\
Low-frequency fraction & 4 & +0.00 & -0.60 & +0.20 & +0.46 & 0/4\\
Mid-frequency fraction & 4 & -0.80 & +0.00 & -0.80 & +0.15 & 1/4\\
High-frequency fraction & 4 & +0.00 & +0.00 & -0.20 & -0.45 & 4/4\\
Spatial spectral entropy & 4 & +0.00 & +0.60 & -0.40 & -0.47 & 4/4\\
Spectral centroid & 4 & +0.00 & +0.00 & +0.40 & -0.49 & 4/4\\
Spectral bandwidth & 4 & +0.80 & +0.60 & +0.40 & -0.31 & 4/4\\
Single-frame IDM & 5 & +0.40 & -0.70 & -0.30 & -0.75 & 4/5\\
\bottomrule\end{tabular*}\end{table}

\begin{table}[!htp]\centering\footnotesize
\caption{\textbf{Selecting five candidates.} All frozen encoders with success and probe measurements are eligible. ``Best'' is the highest success retained by the shortlist; ``Full'' is the highest in the full cohort; ``Random'' is the exact expected best of five uniformly sampled candidates. All scores are success percentages; regret is in points.}
\label{tab:screening}\setlength{\tabcolsep}{5pt}
\begin{tabular*}{\linewidth}{@{\extracolsep{\fill}}llrrrr@{}}\toprule
Setting & Selection readout & Best & Full & Regret & Random\\\midrule
Casa ST & Two-frame IDM & 32.61 & 43.17 & 10.56 & 36.32\\
Casa ST & Robot state & 32.61 & 43.17 & 10.56 & 36.32\\
Casa ST & Single-frame IDM & 32.50 & 43.17 & 10.67 & 36.32\\
Casa MT & Two-frame IDM & 31.22 & 35.39 & 4.17 & 32.17\\
Casa MT & Robot state & 34.78 & 35.39 & 0.61 & 32.17\\
Casa MT & Single-frame IDM & 34.78 & 35.39 & 0.61 & 32.17\\
Twin MT & Two-frame IDM & 53.22 & 53.67 & 0.45 & 50.52\\
Twin MT & Robot state & 53.22 & 53.67 & 0.45 & 50.52\\
Twin MT & Single-frame IDM & 46.78 & 53.67 & 6.89 & 50.52\\
\bottomrule\end{tabular*}\end{table}

\paragraph{Interpreting associations and screening.}
The comparison counts policy-training runs, not total wall time, and includes VLMs. In Casa ST, IDM-5 retains 32.61\% against 36.32\% for random-five and 43.17\% for the full cohort; Casa MT also falls below random selection. On Twin, IDM-5 retains 53.22\% against 50.52\% for random-five, but IDM-3 falls below random-three (45.17\% versus 47.33\%). These retrospective results describe benchmark- and budget-dependent associations, not a generally reliable selection rule. All corrected metrics in the primary diagnostic export enter the within-cohort correction; none is filtered by its observed correlation or $p$-value. These $q$-values apply to the stated 24/25-metric families, not the larger legacy collection. Vision scores and adaptation analyses form separate comparisons. Correlation, conditional significance, and candidate-selection quality answer different questions; none alone establishes a transferable selection rule.

\clearpage
\subsection{Published vision benchmarks}
\label{app:vision}
Table~\ref{tab:app-vision-corr} uses exact checkpoint matches and one protocol per metric. ImageNet linear/kNN, ADE20k, Cityscapes, VOC, NYUv2, and KITTI use the LingBot-Vision evaluation tables~\citep{lingbot}; NAVI and SPair use the DINOv3 comparison with C-RADIOv4-H added from its corresponding evaluation~\citep{dinov3,radio}; DAVIS uses the medium-resolution DINOv3 label-propagation comparison. Depth is an error, so negative $\rho$ is favorable. The DINOv2/DINOv3 Small, Base, and Large comparisons share model size but retain their released preprocessing and patch interfaces.

\begin{table}[!ht]\centering\small
\caption{\textbf{Published vision score versus frozen MT success.} Raw Spearman correlation and two-sided permutation $p$; all ten primary vision metrics have $p>0.05$ on both suites.}
\label{tab:app-vision-corr}\setlength{\tabcolsep}{8pt}
\begin{tabular*}{\linewidth}{@{\extracolsep{\fill}}lr rr rr@{}}\toprule
& & \multicolumn{2}{c}{RoboCasa} & \multicolumn{2}{c}{RoboTwin}\\
Metric & $n$ & $\rho$ & $p$ & $\rho$ & $p$\\\midrule
ImageNet linear & 9 & +0.07 & 0.880 & +0.38 & 0.316\\
ImageNet kNN & 9 & +0.02 & 0.982 & +0.35 & 0.363\\
ADE20k & 9 & +0.10 & 0.810 & -0.02 & 0.982\\
Cityscapes & 9 & +0.10 & 0.810 & -0.02 & 0.982\\
VOC & 9 & +0.10 & 0.809 & -0.43 & 0.249\\
NYUv2 & 9 & +0.00 & 1.000 & +0.18 & 0.644\\
KITTI & 9 & +0.00 & 1.000 & +0.00 & 1.000\\
NAVI & 10 & +0.08 & 0.837 & +0.10 & 0.785\\
SPair & 10 & +0.30 & 0.406 & -0.01 & 1.000\\
DAVIS medium & 9 & -0.27 & 0.482 & -0.29 & 0.443\\
\bottomrule\end{tabular*}\end{table}

Additional same-protocol blocks are analyzed separately in Table~\ref{tab:vision-secondary}, including eight-frame video classification within DINOv2. These size sweeps test scaling within a family, rather than broad encoder selection. Scores from different probing heads, resolutions, or fine-tuning protocols are not pooled. RADIOv2.5-H/g scores do not stand in for the default Base checkpoint; C-RADIOv4-SO400M scores do not stand in for H. The video attentive-probe and YouTube-VOS/MOSE blocks lack a comparable cohort of three exact policy checkpoints under one protocol and resolution. Isolated scores remain descriptive.

\begin{table}[!ht]\centering\footnotesize
\caption{\textbf{Additional published-score associations, separated by protocol.} D2 and D3 denote the DINOv2 and DINOv3 evaluation protocols. D3 dense uses 1,024 patch tokens. Correlations are raw; smaller depth errors are favorable.}
\label{tab:vision-secondary}\setlength{\tabcolsep}{4pt}
\begin{tabular*}{\linewidth}{@{\extracolsep{\fill}}llr rr rr@{}}\toprule
& & & \multicolumn{2}{c}{RoboCasa} & \multicolumn{2}{c}{RoboTwin}\\
Protocol & Metric & $n$ & $\rho$ & $p$ & $\rho$ & $p$\\\midrule
D2 video & UCF101 & 4 & +0.40 & 0.750 & +0.80 & 0.333\\
D2 video & SSv2 & 4 & +0.40 & 0.750 & +0.80 & 0.333\\
D2 video & K400 & 4 & +0.40 & 0.750 & +0.80 & 0.333\\
D2 classification & ImageNet val & 4 & +0.40 & 0.750 & +0.80 & 0.333\\
D2 classification & ImageNet V2 & 4 & +0.40 & 0.750 & +0.80 & 0.333\\
D2 classification & ImageNet ReaL & 4 & +0.40 & 0.750 & +0.80 & 0.333\\
D2 classification & ImageNet kNN & 4 & +0.32 & 0.667 & +0.95 & 0.167\\
D2 dense & ADE20k & 4 & +0.40 & 0.750 & +0.80 & 0.333\\
D2 dense & Cityscapes & 4 & +0.40 & 0.750 & +0.80 & 0.333\\
D2 dense & VOC & 4 & +0.80 & 0.333 & +0.40 & 0.750\\
D2 dense & NYUv2 & 4 & -0.40 & 0.750 & -0.80 & 0.333\\
D2 dense & KITTI & 4 & -0.40 & 0.750 & -0.80 & 0.333\\
D3 dense & ADE20k & 5 & +0.10 & 0.950 & +0.10 & 0.950\\
D3 dense & NYUv2 & 5 & -0.05 & 1.000 & -0.41 & 0.500\\
\bottomrule\end{tabular*}\end{table}
\clearpage

\section{Complete Results}
\label{app:results}

This section lists the closed-loop results behind Section~\ref{sec:results}. A suite score is the unweighted mean of its task scores. The tables give the underlying task values and evaluation deviations.

\par\medskip\noindent\begin{minipage}{\linewidth}
\centering
\captionof{table}{\textbf{Task identifiers.} RoboCasa-GR1 tasks (C) and RoboTwin~2.0 tasks (W) used in every per-task table. ``PnP'' denotes pick-and-place; the RoboCasa single-task data-scale sweep uses C1--C4.}
\label{tab:app-tasks}
\begingroup
\arrayrulecolor{hiveInk}
\setlength{\tabcolsep}{5pt}
\renewcommand{\arraystretch}{1.1}
\small
\newcommand{\sv}[2]{#1{\tiny\,$\pm$#2}}

\begin{tabular*}{\linewidth}{@{\extracolsep{\fill}}clcl}
\toprule[0.8pt]
\textbf{ID} & \textbf{RoboCasa-GR1 task} & \textbf{ID} & \textbf{RoboTwin~2.0 task}\\
\midrule[0.45pt]
C1 & PnP Bottle To Cabinet Close & W1 & Adjust Bottle\\
C2 & PnP Can To Drawer Close & W2 & Beat Block Hammer\\
C3 & PnP Milk To Microwave Close & W3 & Click Alarmclock\\
C4 & PnP Novel From Cuttingboard To Basket & W4 & Handover Block\\
C5 & PnP Novel From Cuttingboard To Pan & W5 & Move Playingcard Away\\
C6 & PnP Novel From Placemat To Basket & W6 & Open Laptop\\
C7 & PnP Novel From Placemat To Plate & W7 & Place Can Basket\\
C8 & PnP Novel From Plate To Cardboardbox & W8 & Rotate QRcode\\
C9 & PnP Novel From Plate To Plate & W9 & Stamp Seal\\
C10 & PnP Novel From Tray To Plate & W10 & Turn Switch\\
C11 & PnP Novel From Tray To Pot & W11 & Place Burger Fries\\
C12 & PnP Novel From Tray To Tieredbasket & W12 & Lift Pot\\
\bottomrule[0.8pt]
\end{tabular*}
\endgroup
\end{minipage}\par\medskip

\subsection{Task-level success}
Tables~\ref{tab:app-casa-st}, \ref{tab:app-casa-mt}, and~\ref{tab:app-twin} give per-task success for every frozen encoder, every available layer-16 variant, every fine-tuned encoder, the input and query-token ablations, and the global-average-pooling readout discussed in Section~\ref{sec:results-gap}. To keep the page format, each table is split in two: tasks 1--6, then tasks 7--12 with the suite average.

\begin{table}[!p]
\centering
\caption{\textbf{RoboCasa single-task.} Tasks C1--C6; success (\%), with evaluation deviations in smaller type.}
\label{tab:app-casa-st}
\begingroup\footnotesize
\setlength{\tabcolsep}{2.4pt}\renewcommand{\arraystretch}{.90}
\begin{tabular*}{\linewidth}{@{\extracolsep{\fill}}lcccccc@{}}
\toprule
\textbf{Encoder / setting} & \textbf{C1} & \textbf{C2} & \textbf{C3} & \textbf{C4} & \textbf{C5} & \textbf{C6}\\\midrule
VGGT-$\Omega$ & 48.0{\scriptsize\,$\pm$6.5} & 74.0{\scriptsize\,$\pm$4.3} & 50.0{\scriptsize\,$\pm$5.9} & 39.3{\scriptsize\,$\pm$2.5} & 59.3{\scriptsize\,$\pm$2.5} & 32.0{\scriptsize\,$\pm$4.3}\\
LingBot-Vision & 49.3{\scriptsize\,$\pm$3.8} & 68.7{\scriptsize\,$\pm$2.5} & 42.0{\scriptsize\,$\pm$4.9} & 30.0{\scriptsize\,$\pm$3.3} & 38.0{\scriptsize\,$\pm$1.6} & 19.3{\scriptsize\,$\pm$0.9}\\
DINOv2 & 60.7{\scriptsize\,$\pm$5.0} & 61.3{\scriptsize\,$\pm$3.4} & 49.3{\scriptsize\,$\pm$6.2} & 26.7{\scriptsize\,$\pm$0.9} & 42.7{\scriptsize\,$\pm$3.4} & 16.0{\scriptsize\,$\pm$1.6}\\
DINOv3 & 53.3{\scriptsize\,$\pm$2.5} & 60.0{\scriptsize\,$\pm$5.7} & 47.3{\scriptsize\,$\pm$6.6} & 17.3{\scriptsize\,$\pm$5.7} & 33.3{\scriptsize\,$\pm$7.4} & 15.3{\scriptsize\,$\pm$3.4}\\
MAE & 55.3{\scriptsize\,$\pm$6.2} & 64.7{\scriptsize\,$\pm$0.9} & 40.7{\scriptsize\,$\pm$4.1} & 14.0{\scriptsize\,$\pm$1.6} & 5.3{\scriptsize\,$\pm$5.0} & 1.3{\scriptsize\,$\pm$0.9}\\
V-JEPA~2.1 & 44.0{\scriptsize\,$\pm$4.3} & 54.7{\scriptsize\,$\pm$0.9} & 34.7{\scriptsize\,$\pm$4.1} & 1.3{\scriptsize\,$\pm$0.9} & 30.7{\scriptsize\,$\pm$3.4} & 9.3{\scriptsize\,$\pm$3.4}\\
Xiaomi-Robotics-1 & 40.7{\scriptsize\,$\pm$2.5} & 63.3{\scriptsize\,$\pm$8.1} & 44.0{\scriptsize\,$\pm$8.6} & 14.0{\scriptsize\,$\pm$2.8} & 17.3{\scriptsize\,$\pm$3.4} & 5.3{\scriptsize\,$\pm$2.5}\\
\bottomrule\end{tabular*}\endgroup
\vspace{3pt}
\centering
\caption{\textbf{RoboCasa single-task (continued).} Tasks C7--C12. Avg. is the twelve-task mean; uncertainty follows Section~\ref{sec:settings}.}
\label{tab:app-casa-st-b}
\begingroup\footnotesize
\setlength{\tabcolsep}{2.4pt}\renewcommand{\arraystretch}{.90}
\begin{tabular*}{\linewidth}{@{\extracolsep{\fill}}lccccccc@{}}
\toprule
\textbf{Encoder / setting} & \textbf{C7} & \textbf{C8} & \textbf{C9} & \textbf{C10} & \textbf{C11} & \textbf{C12} & \textbf{Avg.}\\\midrule
VGGT-$\Omega$ & 14.7{\scriptsize\,$\pm$1.9} & 14.0{\scriptsize\,$\pm$2.8} & 57.3{\scriptsize\,$\pm$6.2} & 49.3{\scriptsize\,$\pm$6.8} & 52.0{\scriptsize\,$\pm$3.3} & 28.0{\scriptsize\,$\pm$7.5} & 43.2{\scriptsize\,$\pm$1.4}\\
LingBot-Vision & 10.7{\scriptsize\,$\pm$2.5} & 9.3{\scriptsize\,$\pm$5.7} & 38.0{\scriptsize\,$\pm$1.6} & 32.0{\scriptsize\,$\pm$3.3} & 30.0{\scriptsize\,$\pm$0.0} & 14.0{\scriptsize\,$\pm$3.3} & 31.8{\scriptsize\,$\pm$0.9}\\
DINOv2 & 14.0{\scriptsize\,$\pm$0.0} & 21.3{\scriptsize\,$\pm$0.9} & 41.3{\scriptsize\,$\pm$3.4} & 51.3{\scriptsize\,$\pm$2.5} & 43.3{\scriptsize\,$\pm$0.9} & 20.0{\scriptsize\,$\pm$4.3} & 37.3{\scriptsize\,$\pm$0.9}\\
DINOv3 & 14.7{\scriptsize\,$\pm$5.2} & 16.7{\scriptsize\,$\pm$3.4} & 32.0{\scriptsize\,$\pm$7.1} & 50.7{\scriptsize\,$\pm$2.5} & 35.3{\scriptsize\,$\pm$1.9} & 14.0{\scriptsize\,$\pm$2.8} & 32.5{\scriptsize\,$\pm$1.4}\\
MAE & 13.3{\scriptsize\,$\pm$6.6} & 3.3{\scriptsize\,$\pm$2.5} & 11.3{\scriptsize\,$\pm$6.6} & 33.3{\scriptsize\,$\pm$0.9} & 10.7{\scriptsize\,$\pm$1.9} & 21.3{\scriptsize\,$\pm$0.9} & 22.9{\scriptsize\,$\pm$1.1}\\
V-JEPA~2.1 & 9.3{\scriptsize\,$\pm$3.4} & 8.0{\scriptsize\,$\pm$5.9} & 28.7{\scriptsize\,$\pm$5.2} & 28.0{\scriptsize\,$\pm$1.6} & 19.3{\scriptsize\,$\pm$3.8} & 12.7{\scriptsize\,$\pm$1.9} & 23.4{\scriptsize\,$\pm$1.0}\\
Xiaomi-Robotics-1 & 14.0{\scriptsize\,$\pm$2.8} & 9.3{\scriptsize\,$\pm$2.5} & 28.7{\scriptsize\,$\pm$8.4} & 28.0{\scriptsize\,$\pm$5.9} & 21.3{\scriptsize\,$\pm$5.0} & 12.0{\scriptsize\,$\pm$5.7} & 24.8{\scriptsize\,$\pm$1.6}\\
\bottomrule\end{tabular*}\endgroup
\end{table}
\begin{table}[!p]
\centering
\caption{\textbf{RoboCasa multi-task.} Tasks C1--C6; success (\%), with evaluation deviations in smaller type.}
\label{tab:app-casa-mt}
\begingroup\footnotesize
\setlength{\tabcolsep}{2.4pt}\renewcommand{\arraystretch}{.90}
\begin{tabular*}{\linewidth}{@{\extracolsep{\fill}}lcccccc@{}}
\toprule
\textbf{Encoder / setting} & \textbf{C1} & \textbf{C2} & \textbf{C3} & \textbf{C4} & \textbf{C5} & \textbf{C6}\\\midrule
VGGT-$\Omega$ & 29.3{\scriptsize\,$\pm$5.7} & 38.0{\scriptsize\,$\pm$1.6} & 34.0{\scriptsize\,$\pm$4.3} & 39.3{\scriptsize\,$\pm$2.5} & 52.0{\scriptsize\,$\pm$1.6} & 32.0{\scriptsize\,$\pm$4.3}\\
LingBot-Vision & 34.0{\scriptsize\,$\pm$8.2} & 49.3{\scriptsize\,$\pm$9.0} & 34.7{\scriptsize\,$\pm$5.7} & 27.3{\scriptsize\,$\pm$3.4} & 44.0{\scriptsize\,$\pm$2.8} & 12.7{\scriptsize\,$\pm$2.5}\\
SPA & 28.0{\scriptsize\,$\pm$2.8} & 28.0{\scriptsize\,$\pm$4.3} & 28.0{\scriptsize\,$\pm$10.2} & 8.0{\scriptsize\,$\pm$1.6} & 20.0{\scriptsize\,$\pm$2.8} & 11.3{\scriptsize\,$\pm$3.4}\\
DINOv2 & 41.3{\scriptsize\,$\pm$5.2} & 37.3{\scriptsize\,$\pm$2.5} & 38.0{\scriptsize\,$\pm$4.3} & 36.7{\scriptsize\,$\pm$4.1} & 49.3{\scriptsize\,$\pm$2.5} & 20.7{\scriptsize\,$\pm$3.4}\\
DINOv3 & 50.0{\scriptsize\,$\pm$4.3} & 34.0{\scriptsize\,$\pm$1.6} & 38.7{\scriptsize\,$\pm$6.2} & 20.7{\scriptsize\,$\pm$5.2} & 41.3{\scriptsize\,$\pm$5.2} & 22.0{\scriptsize\,$\pm$3.3}\\
MAE & 36.7{\scriptsize\,$\pm$7.4} & 39.3{\scriptsize\,$\pm$5.0} & 26.7{\scriptsize\,$\pm$0.9} & 24.0{\scriptsize\,$\pm$1.6} & 40.7{\scriptsize\,$\pm$6.6} & 16.0{\scriptsize\,$\pm$4.3}\\
V-JEPA~2.1 & 26.7{\scriptsize\,$\pm$5.7} & 26.7{\scriptsize\,$\pm$2.5} & 30.7{\scriptsize\,$\pm$7.5} & 14.0{\scriptsize\,$\pm$0.0} & 36.7{\scriptsize\,$\pm$5.0} & 4.7{\scriptsize\,$\pm$2.5}\\
\bottomrule\end{tabular*}\endgroup
\vspace{3pt}
\centering
\caption{\textbf{RoboCasa multi-task (continued).} Tasks C7--C12. Avg. is the twelve-task mean; uncertainty follows Section~\ref{sec:settings}.}
\label{tab:app-casa-mt-b}
\begingroup\footnotesize
\setlength{\tabcolsep}{2.4pt}\renewcommand{\arraystretch}{.90}
\begin{tabular*}{\linewidth}{@{\extracolsep{\fill}}lccccccc@{}}
\toprule
\textbf{Encoder / setting} & \textbf{C7} & \textbf{C8} & \textbf{C9} & \textbf{C10} & \textbf{C11} & \textbf{C12} & \textbf{Avg.}\\\midrule
VGGT-$\Omega$ & 21.3{\scriptsize\,$\pm$2.5} & 16.0{\scriptsize\,$\pm$4.9} & 44.0{\scriptsize\,$\pm$8.6} & 46.0{\scriptsize\,$\pm$3.3} & 45.3{\scriptsize\,$\pm$5.7} & 27.3{\scriptsize\,$\pm$5.0} & 35.4{\scriptsize\,$\pm$1.3}\\
LingBot-Vision & 10.7{\scriptsize\,$\pm$1.9} & 14.7{\scriptsize\,$\pm$3.4} & 34.7{\scriptsize\,$\pm$6.2} & 34.7{\scriptsize\,$\pm$5.2} & 35.3{\scriptsize\,$\pm$2.5} & 20.7{\scriptsize\,$\pm$5.2} & 29.4{\scriptsize\,$\pm$1.5}\\
SPA & 10.0{\scriptsize\,$\pm$2.8} & 2.0{\scriptsize\,$\pm$1.6} & 15.3{\scriptsize\,$\pm$0.9} & 17.3{\scriptsize\,$\pm$2.5} & 19.3{\scriptsize\,$\pm$0.9} & 10.7{\scriptsize\,$\pm$3.8} & 16.5{\scriptsize\,$\pm$1.1}\\
DINOv2 & 28.7{\scriptsize\,$\pm$2.5} & 14.0{\scriptsize\,$\pm$3.3} & 34.0{\scriptsize\,$\pm$2.8} & 46.0{\scriptsize\,$\pm$13.4} & 38.7{\scriptsize\,$\pm$3.4} & 2.7{\scriptsize\,$\pm$0.9} & 32.3{\scriptsize\,$\pm$1.4}\\
DINOv3 & 23.3{\scriptsize\,$\pm$7.4} & 25.3{\scriptsize\,$\pm$0.9} & 26.0{\scriptsize\,$\pm$4.3} & 46.7{\scriptsize\,$\pm$5.2} & 34.7{\scriptsize\,$\pm$2.5} & 12.0{\scriptsize\,$\pm$4.9} & 31.2{\scriptsize\,$\pm$1.3}\\
MAE & 10.7{\scriptsize\,$\pm$6.8} & 8.0{\scriptsize\,$\pm$4.3} & 30.0{\scriptsize\,$\pm$5.7} & 26.7{\scriptsize\,$\pm$5.2} & 29.3{\scriptsize\,$\pm$2.5} & 15.3{\scriptsize\,$\pm$6.6} & 25.3{\scriptsize\,$\pm$1.5}\\
V-JEPA~2.1 & 13.3{\scriptsize\,$\pm$1.9} & 5.3{\scriptsize\,$\pm$3.4} & 16.0{\scriptsize\,$\pm$1.6} & 12.7{\scriptsize\,$\pm$8.2} & 22.0{\scriptsize\,$\pm$1.6} & 15.3{\scriptsize\,$\pm$0.9} & 18.7{\scriptsize\,$\pm$1.2}\\
\bottomrule\end{tabular*}\endgroup
\end{table}
\begin{table}[!p]
\centering
\caption{\textbf{RoboTwin multi-task.} Tasks W1--W6; success (\%), with evaluation deviations in smaller type.}
\label{tab:app-twin}
\begingroup\footnotesize
\setlength{\tabcolsep}{2.4pt}\renewcommand{\arraystretch}{.84}
\begin{tabular*}{\linewidth}{@{\extracolsep{\fill}}lcccccc@{}}
\toprule
\textbf{Encoder / setting} & \textbf{W1} & \textbf{W2} & \textbf{W3} & \textbf{W4} & \textbf{W5} & \textbf{W6}\\\midrule
VGGT-$\Omega$ & 90.0{\scriptsize\,$\pm$4.3} & 74.0{\scriptsize\,$\pm$7.5} & 83.3{\scriptsize\,$\pm$6.8} & 18.0{\scriptsize\,$\pm$1.6} & 44.7{\scriptsize\,$\pm$3.4} & 74.7{\scriptsize\,$\pm$5.7}\\
LingBot-Vision & 91.3{\scriptsize\,$\pm$4.1} & 62.0{\scriptsize\,$\pm$5.9} & 84.7{\scriptsize\,$\pm$3.4} & 26.0{\scriptsize\,$\pm$3.3} & 54.7{\scriptsize\,$\pm$6.8} & 84.0{\scriptsize\,$\pm$8.2}\\
SPA & 19.3{\scriptsize\,$\pm$4.1} & 5.3{\scriptsize\,$\pm$2.5} & 34.0{\scriptsize\,$\pm$4.3} & 0.7{\scriptsize\,$\pm$0.9} & 6.0{\scriptsize\,$\pm$0.0} & 50.7{\scriptsize\,$\pm$1.9}\\
DINOv2 & 96.7{\scriptsize\,$\pm$0.9} & 63.3{\scriptsize\,$\pm$0.9} & 94.7{\scriptsize\,$\pm$0.9} & 14.7{\scriptsize\,$\pm$6.2} & 44.7{\scriptsize\,$\pm$8.4} & 90.7{\scriptsize\,$\pm$3.4}\\
DINOv3 & 72.0{\scriptsize\,$\pm$1.6} & 42.0{\scriptsize\,$\pm$5.9} & 63.3{\scriptsize\,$\pm$6.8} & 15.3{\scriptsize\,$\pm$5.0} & 16.0{\scriptsize\,$\pm$7.1} & 59.3{\scriptsize\,$\pm$2.5}\\
MAE & 77.3{\scriptsize\,$\pm$2.5} & 32.0{\scriptsize\,$\pm$3.3} & 60.7{\scriptsize\,$\pm$10.4} & 0.0{\scriptsize\,$\pm$0.0} & 12.0{\scriptsize\,$\pm$4.3} & 69.3{\scriptsize\,$\pm$1.9}\\
V-JEPA~2.1 & 94.7{\scriptsize\,$\pm$2.5} & 50.7{\scriptsize\,$\pm$1.9} & 92.0{\scriptsize\,$\pm$1.6} & 2.7{\scriptsize\,$\pm$0.9} & 24.7{\scriptsize\,$\pm$5.2} & 80.0{\scriptsize\,$\pm$7.1}\\
\bottomrule\end{tabular*}\endgroup
\vspace{3pt}
\centering
\caption{\textbf{RoboTwin multi-task (continued).} Tasks W7--W12. Avg. is the twelve-task mean; uncertainty follows Section~\ref{sec:settings}.}
\label{tab:app-twin-b}
\begingroup\footnotesize
\setlength{\tabcolsep}{2.4pt}\renewcommand{\arraystretch}{.90}
\begin{tabular*}{\linewidth}{@{\extracolsep{\fill}}lccccccc@{}}
\toprule
\textbf{Encoder / setting} & \textbf{W7} & \textbf{W8} & \textbf{W9} & \textbf{W10} & \textbf{W11} & \textbf{W12} & \textbf{Avg.}\\\midrule
VGGT-$\Omega$ & 48.0{\scriptsize\,$\pm$1.6} & 47.3{\scriptsize\,$\pm$3.8} & 14.0{\scriptsize\,$\pm$4.3} & 57.3{\scriptsize\,$\pm$0.9} & 48.0{\scriptsize\,$\pm$7.1} & 0.7{\scriptsize\,$\pm$0.9} & 50.0{\scriptsize\,$\pm$1.3}\\
LingBot-Vision & 41.3{\scriptsize\,$\pm$3.4} & 40.7{\scriptsize\,$\pm$6.6} & 10.0{\scriptsize\,$\pm$2.8} & 36.0{\scriptsize\,$\pm$1.6} & 30.7{\scriptsize\,$\pm$4.1} & 0.0{\scriptsize\,$\pm$0.0} & 46.8{\scriptsize\,$\pm$1.4}\\
SPA & 2.0{\scriptsize\,$\pm$0.0} & 5.3{\scriptsize\,$\pm$1.9} & 1.3{\scriptsize\,$\pm$1.9} & 22.0{\scriptsize\,$\pm$8.6} & 1.3{\scriptsize\,$\pm$0.9} & 0.0{\scriptsize\,$\pm$0.0} & 12.3{\scriptsize\,$\pm$0.9}\\
DINOv2 & 29.3{\scriptsize\,$\pm$3.4} & 51.3{\scriptsize\,$\pm$1.9} & 18.7{\scriptsize\,$\pm$0.9} & 47.3{\scriptsize\,$\pm$1.9} & 39.3{\scriptsize\,$\pm$2.5} & 2.0{\scriptsize\,$\pm$2.8} & 49.4{\scriptsize\,$\pm$1.0}\\
DINOv3 & 9.3{\scriptsize\,$\pm$0.9} & 25.3{\scriptsize\,$\pm$6.6} & 3.3{\scriptsize\,$\pm$0.9} & 31.3{\scriptsize\,$\pm$7.7} & 26.7{\scriptsize\,$\pm$5.2} & 1.3{\scriptsize\,$\pm$0.9} & 30.4{\scriptsize\,$\pm$1.4}\\
MAE & 5.3{\scriptsize\,$\pm$1.9} & 22.7{\scriptsize\,$\pm$6.6} & 1.3{\scriptsize\,$\pm$0.9} & 33.3{\scriptsize\,$\pm$12.3} & 4.7{\scriptsize\,$\pm$1.9} & 1.3{\scriptsize\,$\pm$0.9} & 26.7{\scriptsize\,$\pm$1.6}\\
V-JEPA~2.1 & 8.0{\scriptsize\,$\pm$0.0} & 30.0{\scriptsize\,$\pm$1.6} & 1.3{\scriptsize\,$\pm$0.9} & 53.3{\scriptsize\,$\pm$9.6} & 38.0{\scriptsize\,$\pm$7.1} & 0.0{\scriptsize\,$\pm$0.0} & 39.6{\scriptsize\,$\pm$1.3}\\
\bottomrule\end{tabular*}\endgroup
\end{table}
\clearpage

\subsection{Demonstration and model scale}
Tables~\ref{tab:app-scale-casa-st} to~\ref{tab:app-model-twin-b} give the per-task values behind Figures~\ref{fig:data-scale} and~\ref{fig:m-scale}.

\begin{table}[!htp]
\centering
\caption{\textbf{RoboCasa single-task demonstration sweep.} Success (\%); small type gives the exported evaluation deviation. Task means are averaged over C1--C4. Avg. averages C1--C4; uncertainty conventions are specified in Section~\ref{sec:settings}.}
\label{tab:app-scale-casa-st}
\begingroup\footnotesize
\setlength{\tabcolsep}{2.4pt}\renewcommand{\arraystretch}{1.1}
\begin{tabular*}{\linewidth}{@{\extracolsep{\fill}}lccccc@{}}
\toprule
\textbf{Encoder / setting} & \textbf{C1} & \textbf{C2} & \textbf{C3} & \textbf{C4} & \textbf{Avg.}\\\midrule
DINOv2 / 50 demos & 15.3{\scriptsize\,$\pm$5.2} & 24.0{\scriptsize\,$\pm$1.6} & 21.3{\scriptsize\,$\pm$6.6} & 3.3{\scriptsize\,$\pm$1.9} & 16.0{\scriptsize\,$\pm$2.2}\\
DINOv2 / 100 demos & 36.7{\scriptsize\,$\pm$6.6} & 40.7{\scriptsize\,$\pm$6.8} & 24.7{\scriptsize\,$\pm$0.9} & 6.0{\scriptsize\,$\pm$2.8} & 27.0{\scriptsize\,$\pm$2.5}\\
DINOv2 / 200 demos & 50.7{\scriptsize\,$\pm$11.5} & 55.3{\scriptsize\,$\pm$7.4} & 40.0{\scriptsize\,$\pm$10.2} & 14.0{\scriptsize\,$\pm$1.6} & 40.0{\scriptsize\,$\pm$4.3}\\
DINOv2 / 500 demos & 57.3{\scriptsize\,$\pm$5.2} & 60.0{\scriptsize\,$\pm$5.7} & 44.0{\scriptsize\,$\pm$1.6} & 21.3{\scriptsize\,$\pm$2.5} & 45.7{\scriptsize\,$\pm$2.1}\\
DINOv3 / 50 demos & 13.3{\scriptsize\,$\pm$5.0} & 20.0{\scriptsize\,$\pm$1.6} & 12.0{\scriptsize\,$\pm$1.6} & 2.0{\scriptsize\,$\pm$1.6} & 11.8{\scriptsize\,$\pm$1.4}\\
DINOv3 / 100 demos & 38.0{\scriptsize\,$\pm$3.3} & 32.0{\scriptsize\,$\pm$2.8} & 29.3{\scriptsize\,$\pm$6.8} & 2.0{\scriptsize\,$\pm$1.6} & 25.3{\scriptsize\,$\pm$2.0}\\
DINOv3 / 200 demos & 43.3{\scriptsize\,$\pm$5.0} & 43.3{\scriptsize\,$\pm$2.5} & 31.3{\scriptsize\,$\pm$5.2} & 12.0{\scriptsize\,$\pm$3.3} & 32.5{\scriptsize\,$\pm$2.1}\\
\bottomrule\end{tabular*}\endgroup
\end{table}
\begin{table}[!htp]
\centering
\caption{\textbf{RoboCasa multi-task demonstration sweep.} Success (\%); small type gives the exported evaluation deviation. Task IDs follow Table~\ref{tab:app-tasks}.}
\label{tab:app-scale-casa-mt}
\begingroup\footnotesize
\setlength{\tabcolsep}{2.4pt}\renewcommand{\arraystretch}{1.1}
\begin{tabular*}{\linewidth}{@{\extracolsep{\fill}}lcccccc@{}}
\toprule
\textbf{Encoder / setting} & \textbf{C1} & \textbf{C2} & \textbf{C3} & \textbf{C4} & \textbf{C5} & \textbf{C6}\\\midrule
DINOv2 / 50 demos & 8.0{\scriptsize\,$\pm$4.3} & 4.0{\scriptsize\,$\pm$1.6} & 5.3{\scriptsize\,$\pm$3.4} & 6.0{\scriptsize\,$\pm$2.8} & 7.3{\scriptsize\,$\pm$0.9} & 3.3{\scriptsize\,$\pm$2.5}\\
DINOv2 / 100 demos & 7.3{\scriptsize\,$\pm$3.4} & 4.7{\scriptsize\,$\pm$1.9} & 4.0{\scriptsize\,$\pm$1.6} & 12.0{\scriptsize\,$\pm$4.3} & 7.3{\scriptsize\,$\pm$2.5} & 8.0{\scriptsize\,$\pm$3.3}\\
DINOv2 / 200 demos & 14.0{\scriptsize\,$\pm$3.3} & 15.3{\scriptsize\,$\pm$5.2} & 16.7{\scriptsize\,$\pm$0.9} & 16.0{\scriptsize\,$\pm$2.8} & 11.3{\scriptsize\,$\pm$0.9} & 10.0{\scriptsize\,$\pm$0.0}\\
DINOv2 / 500 demos & 24.7{\scriptsize\,$\pm$4.1} & 22.7{\scriptsize\,$\pm$2.5} & 28.7{\scriptsize\,$\pm$6.6} & 25.3{\scriptsize\,$\pm$5.2} & 26.7{\scriptsize\,$\pm$5.0} & 16.0{\scriptsize\,$\pm$4.3}\\
DINOv3 / 50 demos & 2.0{\scriptsize\,$\pm$1.6} & 2.0{\scriptsize\,$\pm$2.8} & 9.3{\scriptsize\,$\pm$2.5} & 4.0{\scriptsize\,$\pm$3.3} & 2.7{\scriptsize\,$\pm$0.9} & 0.7{\scriptsize\,$\pm$0.9}\\
DINOv3 / 100 demos & 11.3{\scriptsize\,$\pm$3.8} & 6.7{\scriptsize\,$\pm$1.9} & 8.0{\scriptsize\,$\pm$0.0} & 2.7{\scriptsize\,$\pm$2.5} & 6.7{\scriptsize\,$\pm$3.4} & 3.3{\scriptsize\,$\pm$3.4}\\
DINOv3 / 200 demos & 8.7{\scriptsize\,$\pm$3.4} & 8.7{\scriptsize\,$\pm$1.9} & 18.0{\scriptsize\,$\pm$4.3} & 12.7{\scriptsize\,$\pm$3.4} & 24.0{\scriptsize\,$\pm$7.5} & 7.3{\scriptsize\,$\pm$0.9}\\
\bottomrule\end{tabular*}\endgroup
\end{table}
\begin{table}[!htp]
\centering
\caption{\textbf{RoboCasa multi-task demonstration sweep.} Success (\%); small type gives the exported evaluation deviation. Task IDs follow Table~\ref{tab:app-tasks}. Avg. averages all twelve tasks across both parts; uncertainty conventions are specified in Section~\ref{sec:settings}.}
\label{tab:app-scale-casa-mt-b}
\begingroup\footnotesize
\setlength{\tabcolsep}{2.4pt}\renewcommand{\arraystretch}{1.1}
\begin{tabular*}{\linewidth}{@{\extracolsep{\fill}}lccccccc@{}}
\toprule
\textbf{Encoder / setting} & \textbf{C7} & \textbf{C8} & \textbf{C9} & \textbf{C10} & \textbf{C11} & \textbf{C12} & \textbf{Avg.}\\\midrule
DINOv2 / 50 demos & 4.7{\scriptsize\,$\pm$3.4} & 2.0{\scriptsize\,$\pm$0.0} & 5.3{\scriptsize\,$\pm$0.9} & 11.3{\scriptsize\,$\pm$2.5} & 7.3{\scriptsize\,$\pm$1.9} & 1.3{\scriptsize\,$\pm$1.9} & 5.5{\scriptsize\,$\pm$0.7}\\
DINOv2 / 100 demos & 8.7{\scriptsize\,$\pm$4.1} & 4.7{\scriptsize\,$\pm$2.5} & 13.3{\scriptsize\,$\pm$4.1} & 11.3{\scriptsize\,$\pm$1.9} & 15.3{\scriptsize\,$\pm$1.9} & 6.0{\scriptsize\,$\pm$2.8} & 8.6{\scriptsize\,$\pm$0.9}\\
DINOv2 / 200 demos & 14.0{\scriptsize\,$\pm$4.3} & 11.3{\scriptsize\,$\pm$0.9} & 18.0{\scriptsize\,$\pm$2.8} & 24.0{\scriptsize\,$\pm$5.7} & 14.0{\scriptsize\,$\pm$6.5} & 10.0{\scriptsize\,$\pm$4.3} & 14.6{\scriptsize\,$\pm$1.1}\\
DINOv2 / 500 demos & 18.0{\scriptsize\,$\pm$6.5} & 6.7{\scriptsize\,$\pm$0.9} & 24.0{\scriptsize\,$\pm$3.3} & 26.0{\scriptsize\,$\pm$3.3} & 22.7{\scriptsize\,$\pm$9.0} & 14.7{\scriptsize\,$\pm$0.9} & 21.3{\scriptsize\,$\pm$1.4}\\
DINOv3 / 50 demos & 2.0{\scriptsize\,$\pm$1.6} & 1.3{\scriptsize\,$\pm$1.9} & 2.7{\scriptsize\,$\pm$0.9} & 6.0{\scriptsize\,$\pm$5.7} & 3.3{\scriptsize\,$\pm$0.9} & 0.7{\scriptsize\,$\pm$0.9} & 3.1{\scriptsize\,$\pm$0.7}\\
DINOv3 / 100 demos & 6.7{\scriptsize\,$\pm$1.9} & 4.0{\scriptsize\,$\pm$0.0} & 9.3{\scriptsize\,$\pm$0.9} & 12.0{\scriptsize\,$\pm$3.3} & 8.0{\scriptsize\,$\pm$1.6} & 4.0{\scriptsize\,$\pm$1.6} & 6.9{\scriptsize\,$\pm$0.7}\\
DINOv3 / 200 demos & 9.3{\scriptsize\,$\pm$5.0} & 8.7{\scriptsize\,$\pm$1.9} & 14.0{\scriptsize\,$\pm$7.1} & 19.3{\scriptsize\,$\pm$6.2} & 14.0{\scriptsize\,$\pm$7.1} & 6.0{\scriptsize\,$\pm$4.3} & 12.6{\scriptsize\,$\pm$1.4}\\
\bottomrule\end{tabular*}\endgroup
\end{table}
\begin{table}[!htp]
\centering
\caption{\textbf{RoboTwin multi-task demonstration sweep.} Success (\%); small type gives the exported evaluation deviation. Task IDs follow Table~\ref{tab:app-tasks}.}
\label{tab:app-scale-twin}
\begingroup\footnotesize
\setlength{\tabcolsep}{2.4pt}\renewcommand{\arraystretch}{1.1}
\begin{tabular*}{\linewidth}{@{\extracolsep{\fill}}lcccccc@{}}
\toprule
\textbf{Encoder / setting} & \textbf{W1} & \textbf{W2} & \textbf{W3} & \textbf{W4} & \textbf{W5} & \textbf{W6}\\\midrule
DINOv2 / 50 demos & 60.7{\scriptsize\,$\pm$5.2} & 10.0{\scriptsize\,$\pm$1.6} & 56.7{\scriptsize\,$\pm$4.1} & 0.0{\scriptsize\,$\pm$0.0} & 4.0{\scriptsize\,$\pm$2.8} & 48.7{\scriptsize\,$\pm$3.4}\\
DINOv2 / 100 demos & 56.7{\scriptsize\,$\pm$5.2} & 22.0{\scriptsize\,$\pm$2.8} & 72.0{\scriptsize\,$\pm$2.8} & 0.7{\scriptsize\,$\pm$0.9} & 14.7{\scriptsize\,$\pm$3.8} & 70.0{\scriptsize\,$\pm$1.6}\\
DINOv2 / 200 demos & 86.0{\scriptsize\,$\pm$4.9} & 35.3{\scriptsize\,$\pm$4.1} & 73.3{\scriptsize\,$\pm$7.4} & 0.0{\scriptsize\,$\pm$0.0} & 16.0{\scriptsize\,$\pm$4.3} & 72.0{\scriptsize\,$\pm$1.6}\\
DINOv3 / 50 demos & 52.0{\scriptsize\,$\pm$1.6} & 13.3{\scriptsize\,$\pm$2.5} & 34.7{\scriptsize\,$\pm$1.9} & 0.7{\scriptsize\,$\pm$0.9} & 3.3{\scriptsize\,$\pm$2.5} & 44.7{\scriptsize\,$\pm$4.1}\\
DINOv3 / 100 demos & 58.7{\scriptsize\,$\pm$3.4} & 26.7{\scriptsize\,$\pm$0.9} & 50.7{\scriptsize\,$\pm$5.0} & 6.0{\scriptsize\,$\pm$0.0} & 6.7{\scriptsize\,$\pm$2.5} & 54.7{\scriptsize\,$\pm$4.1}\\
DINOv3 / 200 demos & 60.7{\scriptsize\,$\pm$2.5} & 37.3{\scriptsize\,$\pm$4.1} & 53.3{\scriptsize\,$\pm$1.9} & 10.0{\scriptsize\,$\pm$1.6} & 8.7{\scriptsize\,$\pm$0.9} & 54.7{\scriptsize\,$\pm$7.4}\\
V-JEPA~2.1 / 50 demos & 34.0{\scriptsize\,$\pm$3.3} & 5.3{\scriptsize\,$\pm$0.9} & 28.0{\scriptsize\,$\pm$8.6} & 0.0{\scriptsize\,$\pm$0.0} & 4.0{\scriptsize\,$\pm$4.3} & 33.3{\scriptsize\,$\pm$4.1}\\
\bottomrule\end{tabular*}\endgroup
\end{table}
\begin{table}[!htp]
\centering
\caption{\textbf{RoboTwin multi-task demonstration sweep.} Success (\%); small type gives the exported evaluation deviation. Task IDs follow Table~\ref{tab:app-tasks}. Avg. averages all twelve tasks across both parts; uncertainty conventions are specified in Section~\ref{sec:settings}.}
\label{tab:app-scale-twin-b}
\begingroup\footnotesize
\setlength{\tabcolsep}{2.4pt}\renewcommand{\arraystretch}{1.1}
\begin{tabular*}{\linewidth}{@{\extracolsep{\fill}}lccccccc@{}}
\toprule
\textbf{Encoder / setting} & \textbf{W7} & \textbf{W8} & \textbf{W9} & \textbf{W10} & \textbf{W11} & \textbf{W12} & \textbf{Avg.}\\\midrule
DINOv2 / 50 demos & 0.7{\scriptsize\,$\pm$0.9} & 4.0{\scriptsize\,$\pm$0.0} & 0.0{\scriptsize\,$\pm$0.0} & 22.7{\scriptsize\,$\pm$9.0} & 10.7{\scriptsize\,$\pm$3.4} & 0.0{\scriptsize\,$\pm$0.0} & 18.2{\scriptsize\,$\pm$1.1}\\
DINOv2 / 100 demos & 2.7{\scriptsize\,$\pm$0.9} & 12.0{\scriptsize\,$\pm$6.5} & 2.7{\scriptsize\,$\pm$2.5} & 37.3{\scriptsize\,$\pm$3.8} & 10.0{\scriptsize\,$\pm$2.8} & 0.0{\scriptsize\,$\pm$0.0} & 25.1{\scriptsize\,$\pm$1.0}\\
DINOv2 / 200 demos & 4.0{\scriptsize\,$\pm$3.3} & 22.7{\scriptsize\,$\pm$4.1} & 6.7{\scriptsize\,$\pm$1.9} & 48.0{\scriptsize\,$\pm$3.3} & 18.0{\scriptsize\,$\pm$3.3} & 0.7{\scriptsize\,$\pm$0.9} & 31.9{\scriptsize\,$\pm$1.1}\\
DINOv3 / 50 demos & 0.0{\scriptsize\,$\pm$0.0} & 6.0{\scriptsize\,$\pm$4.3} & 0.0{\scriptsize\,$\pm$0.0} & 12.7{\scriptsize\,$\pm$2.5} & 4.7{\scriptsize\,$\pm$2.5} & 0.0{\scriptsize\,$\pm$0.0} & 14.3{\scriptsize\,$\pm$0.7}\\
DINOv3 / 100 demos & 2.0{\scriptsize\,$\pm$1.6} & 9.3{\scriptsize\,$\pm$3.4} & 0.0{\scriptsize\,$\pm$0.0} & 10.7{\scriptsize\,$\pm$1.9} & 9.3{\scriptsize\,$\pm$0.9} & 0.0{\scriptsize\,$\pm$0.0} & 19.6{\scriptsize\,$\pm$0.7}\\
DINOv3 / 200 demos & 6.7{\scriptsize\,$\pm$1.9} & 27.3{\scriptsize\,$\pm$3.4} & 0.7{\scriptsize\,$\pm$0.9} & 14.0{\scriptsize\,$\pm$1.6} & 18.7{\scriptsize\,$\pm$4.7} & 0.7{\scriptsize\,$\pm$0.9} & 24.4{\scriptsize\,$\pm$0.9}\\
V-JEPA~2.1 / 50 demos & 1.3{\scriptsize\,$\pm$0.9} & 1.3{\scriptsize\,$\pm$1.9} & 0.0{\scriptsize\,$\pm$0.0} & 11.3{\scriptsize\,$\pm$1.9} & 3.3{\scriptsize\,$\pm$0.9} & 0.7{\scriptsize\,$\pm$0.9} & 10.2{\scriptsize\,$\pm$1.0}\\
\bottomrule\end{tabular*}\endgroup
\end{table}
\begin{table}[!htp]
\centering
\caption{\textbf{RoboCasa model-size sweep.} Success (\%); small type gives the exported evaluation deviation. Task IDs follow Table~\ref{tab:app-tasks}.}
\label{tab:app-model-casa}
\begingroup\footnotesize
\setlength{\tabcolsep}{2.4pt}\renewcommand{\arraystretch}{1.1}
\begin{tabular*}{\linewidth}{@{\extracolsep{\fill}}lcccccc@{}}
\toprule
\textbf{Encoder / setting} & \textbf{C1} & \textbf{C2} & \textbf{C3} & \textbf{C4} & \textbf{C5} & \textbf{C6}\\\midrule
DINOv2 S & 32.0{\scriptsize\,$\pm$7.5} & 40.0{\scriptsize\,$\pm$2.8} & 24.0{\scriptsize\,$\pm$1.6} & 25.3{\scriptsize\,$\pm$2.5} & 44.0{\scriptsize\,$\pm$4.3} & 13.3{\scriptsize\,$\pm$7.5}\\
DINOv2 L & 31.3{\scriptsize\,$\pm$4.1} & 44.7{\scriptsize\,$\pm$3.4} & 22.7{\scriptsize\,$\pm$4.1} & 28.0{\scriptsize\,$\pm$7.5} & 50.7{\scriptsize\,$\pm$11.5} & 29.3{\scriptsize\,$\pm$6.6}\\
DINOv2 g & 35.3{\scriptsize\,$\pm$7.7} & 36.0{\scriptsize\,$\pm$5.9} & 34.0{\scriptsize\,$\pm$7.5} & 30.7{\scriptsize\,$\pm$3.4} & 42.7{\scriptsize\,$\pm$2.5} & 21.3{\scriptsize\,$\pm$2.5}\\
DINOv3 S & 26.0{\scriptsize\,$\pm$3.3} & 28.7{\scriptsize\,$\pm$7.7} & 17.3{\scriptsize\,$\pm$1.9} & 14.0{\scriptsize\,$\pm$1.6} & 33.3{\scriptsize\,$\pm$4.1} & 9.3{\scriptsize\,$\pm$5.0}\\
DINOv3 L & 47.3{\scriptsize\,$\pm$6.6} & 28.7{\scriptsize\,$\pm$6.8} & 35.3{\scriptsize\,$\pm$5.0} & 21.3{\scriptsize\,$\pm$3.8} & 32.7{\scriptsize\,$\pm$3.8} & 15.3{\scriptsize\,$\pm$4.1}\\
DINOv3 H+ & 44.7{\scriptsize\,$\pm$4.1} & 23.3{\scriptsize\,$\pm$1.9} & 24.0{\scriptsize\,$\pm$1.6} & 24.0{\scriptsize\,$\pm$3.3} & 32.0{\scriptsize\,$\pm$1.6} & 18.7{\scriptsize\,$\pm$3.4}\\
V-JEPA~2.1 L & 40.7{\scriptsize\,$\pm$9.3} & 44.0{\scriptsize\,$\pm$2.8} & 34.0{\scriptsize\,$\pm$3.3} & 38.0{\scriptsize\,$\pm$2.8} & 53.3{\scriptsize\,$\pm$7.4} & 32.0{\scriptsize\,$\pm$7.5}\\
V-JEPA~2.1 g & 44.0{\scriptsize\,$\pm$13.0} & 46.0{\scriptsize\,$\pm$4.9} & 37.3{\scriptsize\,$\pm$2.5} & 36.7{\scriptsize\,$\pm$3.8} & 56.0{\scriptsize\,$\pm$2.8} & 21.3{\scriptsize\,$\pm$1.9}\\
VC-1 L & 40.0{\scriptsize\,$\pm$4.3} & 57.3{\scriptsize\,$\pm$8.1} & 31.3{\scriptsize\,$\pm$2.5} & 27.3{\scriptsize\,$\pm$6.6} & 44.0{\scriptsize\,$\pm$7.1} & 22.0{\scriptsize\,$\pm$1.6}\\
\bottomrule\end{tabular*}\endgroup
\end{table}
\begin{table}[!htp]
\centering
\caption{\textbf{RoboCasa model-size sweep.} Success (\%); small type gives the exported evaluation deviation. Task IDs follow Table~\ref{tab:app-tasks}. Avg. averages all twelve tasks across both parts; uncertainty conventions are specified in Section~\ref{sec:settings}.}
\label{tab:app-model-casa-b}
\begingroup\footnotesize
\setlength{\tabcolsep}{2.4pt}\renewcommand{\arraystretch}{1.1}
\begin{tabular*}{\linewidth}{@{\extracolsep{\fill}}lccccccc@{}}
\toprule
\textbf{Encoder / setting} & \textbf{C7} & \textbf{C8} & \textbf{C9} & \textbf{C10} & \textbf{C11} & \textbf{C12} & \textbf{Avg.}\\\midrule
DINOv2 S & 14.7{\scriptsize\,$\pm$3.4} & 16.7{\scriptsize\,$\pm$9.8} & 30.0{\scriptsize\,$\pm$4.3} & 27.3{\scriptsize\,$\pm$3.4} & 42.0{\scriptsize\,$\pm$4.9} & 19.3{\scriptsize\,$\pm$2.5} & 27.4{\scriptsize\,$\pm$1.5}\\
DINOv2 L & 14.0{\scriptsize\,$\pm$5.9} & 9.3{\scriptsize\,$\pm$0.9} & 30.0{\scriptsize\,$\pm$8.5} & 30.0{\scriptsize\,$\pm$5.9} & 24.7{\scriptsize\,$\pm$3.4} & 20.7{\scriptsize\,$\pm$8.2} & 27.9{\scriptsize\,$\pm$1.9}\\
DINOv2 g & 19.3{\scriptsize\,$\pm$4.1} & 12.0{\scriptsize\,$\pm$0.0} & 33.3{\scriptsize\,$\pm$9.8} & 26.7{\scriptsize\,$\pm$5.2} & 24.0{\scriptsize\,$\pm$0.0} & 25.3{\scriptsize\,$\pm$1.9} & 28.4{\scriptsize\,$\pm$1.5}\\
DINOv3 S & 12.7{\scriptsize\,$\pm$4.7} & 6.7{\scriptsize\,$\pm$2.5} & 18.0{\scriptsize\,$\pm$3.3} & 27.3{\scriptsize\,$\pm$6.8} & 21.3{\scriptsize\,$\pm$2.5} & 11.3{\scriptsize\,$\pm$2.5} & 18.8{\scriptsize\,$\pm$1.2}\\
DINOv3 L & 13.3{\scriptsize\,$\pm$5.2} & 10.7{\scriptsize\,$\pm$3.8} & 21.3{\scriptsize\,$\pm$2.5} & 26.0{\scriptsize\,$\pm$5.9} & 22.7{\scriptsize\,$\pm$2.5} & 17.3{\scriptsize\,$\pm$3.4} & 24.3{\scriptsize\,$\pm$1.3}\\
DINOv3 H+ & 16.7{\scriptsize\,$\pm$1.9} & 14.7{\scriptsize\,$\pm$1.9} & 22.7{\scriptsize\,$\pm$1.9} & 25.3{\scriptsize\,$\pm$7.5} & 31.3{\scriptsize\,$\pm$4.1} & 20.0{\scriptsize\,$\pm$3.3} & 24.8{\scriptsize\,$\pm$1.0}\\
V-JEPA~2.1 L & 17.3{\scriptsize\,$\pm$5.7} & 20.0{\scriptsize\,$\pm$8.2} & 39.3{\scriptsize\,$\pm$9.0} & 43.3{\scriptsize\,$\pm$1.9} & 40.0{\scriptsize\,$\pm$8.6} & 22.0{\scriptsize\,$\pm$4.3} & 35.3{\scriptsize\,$\pm$1.9}\\
\bottomrule\end{tabular*}\endgroup
\end{table}
\begin{table}[!htp]
\centering
\caption{\textbf{RoboTwin model-size sweep.} Success (\%); small type gives the exported evaluation deviation. Task IDs follow Table~\ref{tab:app-tasks}.}
\label{tab:app-model-twin}
\begingroup\footnotesize
\setlength{\tabcolsep}{2.4pt}\renewcommand{\arraystretch}{1.1}
\begin{tabular*}{\linewidth}{@{\extracolsep{\fill}}lcccccc@{}}
\toprule
\textbf{Encoder / setting} & \textbf{W1} & \textbf{W2} & \textbf{W3} & \textbf{W4} & \textbf{W5} & \textbf{W6}\\\midrule
DINOv2 S & 84.0{\scriptsize\,$\pm$8.2} & 32.7{\scriptsize\,$\pm$4.1} & 90.0{\scriptsize\,$\pm$0.0} & 18.7{\scriptsize\,$\pm$6.2} & 46.0{\scriptsize\,$\pm$7.1} & 82.0{\scriptsize\,$\pm$3.3}\\
DINOv2 L & 92.0{\scriptsize\,$\pm$0.0} & 71.3{\scriptsize\,$\pm$3.4} & 93.3{\scriptsize\,$\pm$1.9} & 28.7{\scriptsize\,$\pm$0.9} & 33.3{\scriptsize\,$\pm$2.5} & 96.0{\scriptsize\,$\pm$2.8}\\
DINOv2 g & 80.7{\scriptsize\,$\pm$2.5} & 60.0{\scriptsize\,$\pm$3.3} & 93.3{\scriptsize\,$\pm$0.9} & 34.0{\scriptsize\,$\pm$4.3} & 34.7{\scriptsize\,$\pm$2.5} & 90.7{\scriptsize\,$\pm$0.9}\\
DINOv3 S & 52.7{\scriptsize\,$\pm$8.2} & 31.3{\scriptsize\,$\pm$3.8} & 46.0{\scriptsize\,$\pm$6.5} & 8.0{\scriptsize\,$\pm$1.6} & 21.3{\scriptsize\,$\pm$6.6} & 54.7{\scriptsize\,$\pm$2.5}\\
DINOv3 L & 73.3{\scriptsize\,$\pm$6.2} & 29.3{\scriptsize\,$\pm$5.0} & 58.7{\scriptsize\,$\pm$2.5} & 15.3{\scriptsize\,$\pm$5.2} & 21.3{\scriptsize\,$\pm$0.9} & 63.3{\scriptsize\,$\pm$5.0}\\
DINOv3 H+ & 75.3{\scriptsize\,$\pm$1.9} & 40.0{\scriptsize\,$\pm$7.5} & 72.0{\scriptsize\,$\pm$7.5} & 20.7{\scriptsize\,$\pm$5.0} & 22.7{\scriptsize\,$\pm$5.2} & 50.7{\scriptsize\,$\pm$5.0}\\
V-JEPA~2.1 L & 73.3{\scriptsize\,$\pm$2.5} & 12.0{\scriptsize\,$\pm$0.0} & 34.7{\scriptsize\,$\pm$0.9} & 4.7{\scriptsize\,$\pm$3.4} & 40.7{\scriptsize\,$\pm$3.8} & 84.0{\scriptsize\,$\pm$7.5}\\
V-JEPA~2.1 g & 97.3{\scriptsize\,$\pm$1.9} & 41.3{\scriptsize\,$\pm$3.8} & 96.7{\scriptsize\,$\pm$0.9} & 5.3{\scriptsize\,$\pm$0.9} & 67.3{\scriptsize\,$\pm$5.0} & 96.7{\scriptsize\,$\pm$2.5}\\
VC-1 L & 77.3{\scriptsize\,$\pm$6.2} & 25.3{\scriptsize\,$\pm$1.9} & 64.7{\scriptsize\,$\pm$0.9} & 0.0{\scriptsize\,$\pm$0.0} & 24.7{\scriptsize\,$\pm$8.2} & 74.0{\scriptsize\,$\pm$3.3}\\
\bottomrule\end{tabular*}\endgroup
\end{table}
\begin{table}[!htp]
\centering
\caption{\textbf{RoboTwin model-size sweep.} Success (\%); small type gives the exported evaluation deviation. Task IDs follow Table~\ref{tab:app-tasks}. Avg. averages all twelve tasks across both parts; uncertainty conventions are specified in Section~\ref{sec:settings}.}
\label{tab:app-model-twin-b}
\begingroup\footnotesize
\setlength{\tabcolsep}{2.4pt}\renewcommand{\arraystretch}{1.1}
\begin{tabular*}{\linewidth}{@{\extracolsep{\fill}}lccccccc@{}}
\toprule
\textbf{Encoder / setting} & \textbf{W7} & \textbf{W8} & \textbf{W9} & \textbf{W10} & \textbf{W11} & \textbf{W12} & \textbf{Avg.}\\\midrule
DINOv2 S & 18.7{\scriptsize\,$\pm$5.7} & 45.3{\scriptsize\,$\pm$7.4} & 9.3{\scriptsize\,$\pm$2.5} & 50.0{\scriptsize\,$\pm$5.9} & 39.3{\scriptsize\,$\pm$11.1} & 6.0{\scriptsize\,$\pm$1.6} & 43.5{\scriptsize\,$\pm$1.7}\\
DINOv2 L & 34.7{\scriptsize\,$\pm$6.6} & 49.3{\scriptsize\,$\pm$4.1} & 18.0{\scriptsize\,$\pm$5.7} & 34.0{\scriptsize\,$\pm$9.1} & 62.0{\scriptsize\,$\pm$1.6} & 0.7{\scriptsize\,$\pm$0.9} & 51.1{\scriptsize\,$\pm$1.2}\\
DINOv2 g & 22.0{\scriptsize\,$\pm$4.9} & 50.0{\scriptsize\,$\pm$3.3} & 15.3{\scriptsize\,$\pm$5.2} & 59.3{\scriptsize\,$\pm$2.5} & 64.0{\scriptsize\,$\pm$4.3} & 0.0{\scriptsize\,$\pm$0.0} & 50.3{\scriptsize\,$\pm$0.9}\\
DINOv3 S & 12.0{\scriptsize\,$\pm$1.6} & 16.7{\scriptsize\,$\pm$0.9} & 0.0{\scriptsize\,$\pm$0.0} & 29.3{\scriptsize\,$\pm$2.5} & 22.0{\scriptsize\,$\pm$1.6} & 0.7{\scriptsize\,$\pm$0.9} & 24.6{\scriptsize\,$\pm$1.1}\\
DINOv3 L & 14.0{\scriptsize\,$\pm$4.9} & 28.7{\scriptsize\,$\pm$2.5} & 2.0{\scriptsize\,$\pm$1.6} & 26.0{\scriptsize\,$\pm$2.8} & 13.3{\scriptsize\,$\pm$3.4} & 0.0{\scriptsize\,$\pm$0.0} & 28.8{\scriptsize\,$\pm$1.1}\\
DINOv3 H+ & 15.3{\scriptsize\,$\pm$4.7} & 26.7{\scriptsize\,$\pm$4.1} & 0.7{\scriptsize\,$\pm$0.9} & 27.3{\scriptsize\,$\pm$2.5} & 15.3{\scriptsize\,$\pm$3.4} & 0.7{\scriptsize\,$\pm$0.9} & 30.6{\scriptsize\,$\pm$1.3}\\
V-JEPA~2.1 L & 2.0{\scriptsize\,$\pm$1.6} & 57.3{\scriptsize\,$\pm$7.7} & 0.0{\scriptsize\,$\pm$0.0} & 61.3{\scriptsize\,$\pm$3.4} & 82.0{\scriptsize\,$\pm$1.6} & 0.0{\scriptsize\,$\pm$0.0} & 37.7{\scriptsize\,$\pm$1.1}\\
\bottomrule\end{tabular*}\endgroup
\end{table}
\clearpage

\end{document}